\pdfoutput=1
\documentclass[10pt,twocolumn,letterpaper]{article}

\usepackage[accsupp]{axessibility}
\usepackage[pagenumbers]{cvpr}

\newcommand{\ours}{{SpeeDiff}}
\newcommand{\ourvae}{{ViT-VAE}}
\newcommand{\ourdit}{{Refined-DiT}}
\definecolor{darkgreen}{RGB}{0,100,0}
\definecolor{darkred}{RGB}{139,0,0}

\newcommand{\NA}{-}
\newcommand{\hldigit}[1]{\textbf{#1}}
\newcommand{\hlconf}[1]{\color{violet}{#1}}

\definecolor{diffhl}{HTML}{EEEEEE}

\usepackage{relsize}

\usepackage{amsmath,amsfonts,bm,amssymb,mathtools,amsthm}
\usepackage{color,xcolor,placeins}
\usepackage{thm-restate}

\definecolor{orange}{HTML}{ff6c0c}
\definecolor{blue}{HTML}{1f77b4}
\definecolor{Gray}{gray}{0.85}
\definecolor{LightCyan}{rgb}{0.88,1,1}

\usepackage{url}
\usepackage{xkcdcolors}
\usepackage{mdframed}

\usepackage{microtype}
\usepackage{graphicx}
\usepackage{subcaption}
\usepackage{latexsym}
\usepackage{sidecap}
\usepackage{caption}
\usepackage{booktabs} %
\usepackage{amsfonts}       %
\usepackage{nicefrac}       %
\usepackage{comment}
\usepackage{enumitem}
\usepackage{wrapfig,tikz}
\usepackage{longtable}
\usepackage{bigstrut, tabularx, multirow, makecell, diagbox}
\usepackage[export]{adjustbox}
\usepackage{float}
\usepackage{stackrel}
\usepackage{color}
\usepackage{colortbl}
\usepackage{theoremref}
\usepackage{cases}
\usepackage{stmaryrd}
\usepackage{mathabx}
\usepackage{dsfont}
\usepackage{arydshln}
\usepackage{algorithm}
\usepackage{algpseudocode}

\makeatletter
\usepackage{xspace}
\def\@onedot{\ifx\@let@token.\else.\null\fi\xspace}
\DeclareRobustCommand\onedot{\futurelet\@let@token\@onedot}

\definecolor{blue1}{RGB}{0,128,255}
\definecolor{blue3}{RGB}{0,0,128}
\definecolor{darkpastelgreen}{rgb}{0.01, 0.75, 0.24}
\definecolor{cerulean}{rgb}{0.0, 0.48, 0.65}

\definecolor{darkgreen}{rgb}{0,0.6,0}

\def\eqref#1{equation~(\ref{#1})}

\def\1{\bm{1}}

\def\rvf{{\mathbf{f}}}

\def\rvv{{\mathbf{v}}}

\def\rvx{{\mathbf{x}}}

\def\rvy{{\mathbf{y}}}
\def\rvz{{\mathbf{z}}}

\def\vzero{{\bm{0}}}

\def\vmu{{\bm{\mu}}}
\def\vtheta{{\bm{\theta}}}

\def\vepsilon{{\bm{\epsilon}}}

\def\vomega{{\bm{\omega}}}

\def\vphi{{\bm{\phi}}}
\def\vpsi{{\bm{\psi}}}

\def\mI{{\bm{I}}}

\DeclareMathAlphabet{\mathsfit}{\encodingdefault}{\sfdefault}{m}{sl}
\SetMathAlphabet{\mathsfit}{bold}{\encodingdefault}{\sfdefault}{bx}{n}

\definecolor{cvprblue}{rgb}{0.21,0.49,0.74}
\usepackage[pagebackref,breaklinks,colorlinks,allcolors=cvprblue]{hyperref}

\def\confName{CVPR}
\def\confYear{2026}

\title{SpeeDiff: Scalable Pixel-Anchored End-to-End Latent Diffusion Model}

\author{
  Bingliang Zhang$^{1,2*}$ \qquad Wenda Chu$^{1,2*}$ \qquad Yizhuo Li$^{1}$ \qquad Linjie Yang$^{1}$\\ \,\, Yisong Yue$^{2}$ \qquad Katherine L. Bouman$^{2}$  \qquad Yang Song$^{3}$ \qquad  Qiushan Guo$^{1}$ \\\\\vspace{-12pt}$^{1}$ByteDance Seed\,\,\,\,\,\,\,\,$^{2}$California Institute of Technology\,\,\,\,\,\,\,\,$^{3}$Meta\\
}

\begin{document}

\twocolumn[{%
\renewcommand\twocolumn[1][]{#1}%
\maketitle
\begin{center}
    \centering
    \captionsetup{type=figure}
    \includegraphics[width=\textwidth]{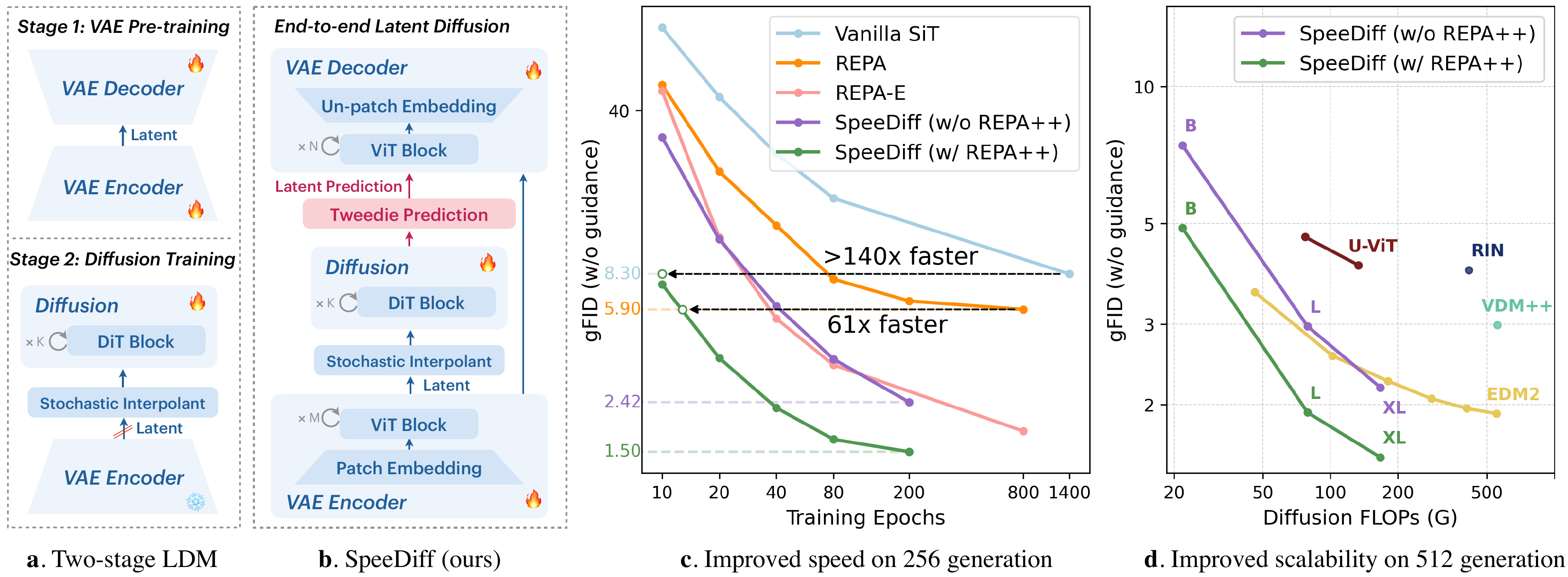}
    \captionof{figure}{\textbf{Overview of {\ours} for end-to-end joint training of VAE and diffusion model from scratch.} (a) Conventional two-stage LDM training: a CNN-based VAE~\cite{vae} is first trained and then frozen, after which a diffusion (usually a DiT~\cite{peebles2023scalable}) is trained on its latent space. (b) {\ours}: a scalable, pixel-anchored end-to-end paradigm that jointly trains the VAE and diffusion model from scratch, without stop-gradient operation, within a fully transformer-based architecture. (c) Training efficiency: on ImageNet 256$\times$256 generation, {\ours} accelerates convergence by over 140$\times$ compared to Vanilla SiT and 61$\times$ compared to REPA. (d) Scalability: on ImageNet 512$\times$512 generation, {\ours} demonstrates clear scalability with a more efficient 32$\times$ compressed VAE with different model size (B, X, XL).
    } 
     \label{fig:teaser}
\end{center}
}]

{\renewcommand{\thefootnote}{\fnsymbol{footnote}}\footnotetext[1]{Work done during internship at ByteDance Seed.}}

\begin{abstract}
We present \textbf{S}calable \textbf{P}ixel-anchored \textbf{E}nd-to-\textbf{e}nd \textbf{Diff}usion (SpeeDiff), a latent diffusion method that jointly trains the VAE and the diffusion model from scratch. In principle, joint training allows the diffusion loss gradient to directly guide the VAE encoder, encouraging the formation of a generation-friendly latent space and potentially yielding faster convergence than the conventional two-stage approach with a pretrained frozen VAE. However, a naive end-to-end implementation severely degrades performance, as unrestricted backpropagation of the diffusion loss leads to latent space collapse. 
Our main technical contribution is a simple yet effective Tweedie Pixel Reconstruction (TPR) loss, which provides additional pixel-level feedback by decoding a predicted clean latent from an intermediate noisy state using Tweedie's formula, thereby alleviating collapse. Furthermore, our method enables jointly scaling a fully transformer-based architecture and enhances representation alignment within the end-to-end framework. 
Our SpeeDiff-XL model achieves over 140$\times$ and 61$\times$ faster training compared to Vanilla SiT and REPA, respectively, while attaining an FID of 1.50 without guidance on ImageNet 256$\times$256 generation. With a more efficient 32$\times$ compressed VAE, our model further reaches an FID of 1.53 without guidance on ImageNet 512$\times$512 generation.

\end{abstract}
\vspace{-10pt}

\section{Introduction}
Latent diffusion models (LDMs)~\cite{vahdat2021scorebased, rombach2022high, peebles2023scalable} have emerged as a dominant paradigm for visual synthesis. These models perform the diffusion process in a compact latent space produced by a visual tokenizer, typically a variational autoencoder (VAE)~\citep{vae}. In the standard two-stage training pipeline, the visual tokenizer is first trained to encode raw visual data into a compressed latent space, after which a diffusion model is learned to operate within that latent domain while keeping the pretrained visual tokenizer frozen.

Recent advances in LDMs have largely focused on enhancing the VAEs by leveraging vision foundation models (VFMs)~\cite{oquab2023dinov2,simeoni2025dinov3,radford2021clip,zhai2023siglip,he2022masked}. Despite their success, VAEs trained with reconstruction  objectives primarily capture low-level pixel statistics, resulting in latent spaces that do not directly associate with sufficient semantic knowledge, posing challenges for the diffusion model to learn efficiently. Motivated by semantically rich representations from self-supervise pretrained VFMs, recent works either incorporate representation alignment to a VFM~\cite{lightdit,leng2025repae} or directly replace the VAE encoder with a VFM~\cite{rae,svg,bi2025vision,chen2025aligning}. These approaches have achieved remarkable improvements in training efficiency and generative quality.

In contrast to improving the visual tokenizer through VFMs, much less attention has been given to enhancing it via end-to-end optimization with direct feedback from the diffusion model training. This idea can in fact be traced back to LSGM~\cite{vahdat2021scorebased}, where the VAE and diffusion model were jointly trained. However, vanilla end-to-end training was found to degrade overall performance, a phenomenon that has also been revisited in recent studies~\cite{leng2025repae}. 
We believe that enabling truly end-to-end optimization holds the potential to further push the performance limits of LDMs, yet existing approaches still lack effective mechanisms to support that.

\begin{figure*}[!t]
    \centering
    \includegraphics[width=\linewidth]{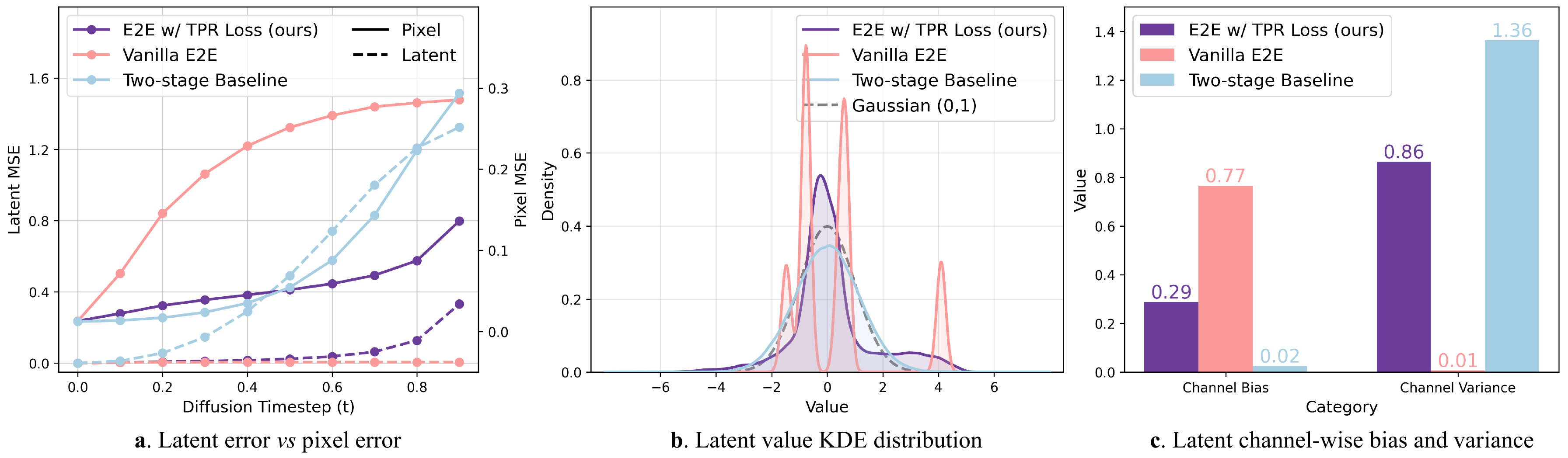}
    \caption{\textbf{Impact of end-to-end training on latent space.} (a) Comparing the denoised latent with its clean counterpart in both latent and pixel space reveals that vanilla end-to-end training collapses the latent representation, while TPR loss preserves meaningful reconstruction signals. (b) KDE visualizations show that vanilla training produces highly peaked, non-Gaussian latent distributions, whereas TPR regularizes the latent space and prevents degeneracy. (c) Channel-wise statistics further indicate that vanilla training induces large per-channel biases and severely suppressed variances, whereas TPR maintains a more balanced and normalized latent representation. Results are computed over 1000 validation images. For (c), \emph{channel bias} is defined as the per-channel absolute mean, and \emph{channel variance} is computed as the per-channel variance across samples and spatial locations.}
    \label{fig:analysis_latent}
    \vspace{-10pt}
\end{figure*}

In this paper, we focus on the core idea of {\hlconf{\emph{enhancing the VAE through end-to-end direct gradient feedback from diffusion training}}}, and show that allowing the latent space to evolve under generative supervision further improves both training efficiency and synthesis quality. After carefully analyzing the vanilla end-to-end training, we find that the diffusion loss tends to collapse the latent space into a degenerate representation. Specifically, we observe that the latent statistics exhibit severely suppressed per-channel variances and large per-channel biass (see \cref{fig:analysis_latent}). We further demonstrate that this latent collapse can be effectively alleviated with a simple \emph{Tweedie Pixel Reconstruction} (TPR) loss, which computes a pixel-level reconstruction loss by decoding the predicted clean latent from an intermediate noisy state using Tweedie’s formula~\cite{efron2011tweedie}. Extensive experiments show that this additional pixel-level feedback anchors the VAE and prevents latent degeneracy, resulting in robust and well-converged end-to-end training.

Building upon the end-to-end training framework, we further improve latent diffusion model through two enhancements. We first refine the architecture by replacing the conventional convolutional VAE (CNN-VAE) with a vision-transformer-based variant (ViT-VAE) and modernizing the diffusion backbone (DiT~\cite{peebles2023scalable,ma2024sit}) with recent architectural improvements. In addition, inspired by REPA~\cite{yu2025repa,leng2025repae}, we introduce \emph{REPA++}, a enhanced representation alignment method in our end-to-end training that aligns both the latent and the intermediate diffusion features with VFM representations (e.g., DINOv3~\cite{simeoni2025dinov3}) and is optimized end-to-end with both diffusion and VAE.

Our final method, {\ours}, is a single-stage end-to-end latent diffusion framework that jointly trains the VAE and diffusion model from scratch without stop-gradient operations (see \cref{fig:teaser}b), simplifying the conventional two-stage pipeline while substantially accelerating diffusion convergence (see \cref{fig:teaser}c) and improving scaling effect (see \cref{fig:teaser}d). 
To assess the contribution of representation alignment, we compare {\ours} variants with and without REPA++. The model without alignment achieves an FID of 2.42, setting a new state-of-the-art among methods that do not rely on VFM alignment, whereas incorporating REPA++ further reduces the FID to 1.50. 
For ImageNet 512$\times$512, {\ours}-XL model using a efficient 32$\times$ compressed VAE achieves an FID of 1.53 without guidance.

\begin{figure*}[!t]
    \centering    \includegraphics[width=\linewidth]{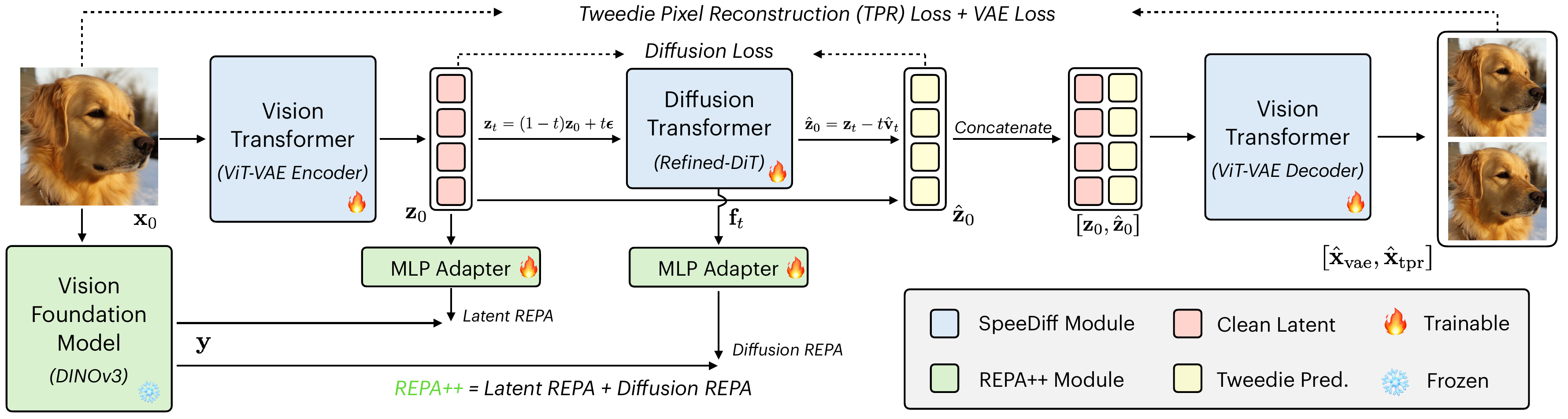 }
    \caption{\textbf{Detailed training pipeline of {\ours}.} The forward pass consists of four major branches: (1) Reconstruction branch: $\mathcal L_{\text{VAE}}$ is computed over reconstructions from the VAE encoder and decoder. (2) Diffusion branch: follows the standard stochastic interpolant and leverages flow prediction as diffusion loss. (3) Tweedie Pixel Reconstruction branch: clean latents are estimated via Tweedie’s formula, decoded, and compared against ground-truth images. (4) REPA++ branch: both latents and intermediate diffusion features are aligned with VFM representations. Notably, {\ours} doesn't requires stop-gradient operation and train both VAE and diffusion from scratch.}
    \label{fig:pipeline}
    \vspace{-10pt}
\end{figure*}

\section{Background}
\paragraph{Variational autoencoder as tokenizer.}
Visual tokenizers employ either continuous~\cite{rombach2022high,labs2025flux1kontextflowmatching}, discrete~\cite{esser2021taming,ma2025unitok}, or unified~\cite{wang2024omnitokenizer,lu2025atoken} latent representations. We focus on continuous tokenizers for visual generation, which are predominantly implemented using variational autoencoders (VAEs)~\cite{vae,rombach2022high}.
Such VAE-based tokenizers have become a fundamental component of modern text-to-image and video generation systems~\cite{labs2025flux1kontextflowmatching,gao2025seedream,polyak2024movie,kong2024hunyuanvideo,wang2024omnitokenizer}.

Formally, the VAE encoder, parameterized by $\vphi$, defines $p_\vphi(\rvz_0 \mid \rvx_0) = \mathcal N(\rvz_0; \vmu_\vphi(\rvx_0), \sigma_\vphi^2(\rvx_0)\mI)$, 
while the decoder $\mathcal D_\vpsi$ reconstruct the image $\hat\rvx_0 = \mathcal D_\vpsi(\rvz_0)$, where $\rvx_0$ and $\rvz_0$ denote the clean image and its latent representation, respectively. They are optimized with reconstruction and variational KL loss given by evidence lower bound (ELBO)~\cite{vae}. To achieve better fidelity, 
the reconstruction term is often complemented with 
perceptual loss~\cite{lpips,fu2023dreamsim} and adversarial loss~\cite{goodfellow2020generative,esser2021taming}.

\vspace{-10pt}
\paragraph{Diffusion models.}
Diffusion and flow-based generative models synthesize images by gradually transforming Gaussian noise into clean data samples~\cite{ddpm,sysde,edm,peebles2023scalable}. However, training these models directly in pixel space remains computationally demanding. Latent diffusion models (LDMs)~\cite{esser2024scaling,rombach2022high,podell2023sdxl,peebles2023scalable} operate in compact latent spaces produced by pretrained tokenizer to reduce costs. In this work, we adopt the LDM formulation under the unified perspective of stochastic interpolants~\cite{albergo2023stochastic,ma2024sit}.

Given a latent embedding $\rvz_0$ and Gaussian noise $\vepsilon \sim \mathcal N(\vzero, \mI)$, the interpolant is defined as
    $\rvz_t = (1-t)\rvz_0 + t\vepsilon,\ t\in [0,1].$
The marginal distribution $p(\rvz_t)$ can be derived by solving the probability flow ODE with a velocity flow, $\rvv(\rvz, t) = \mathbb E[\dot\rvz_t\mid \rvz_t = \rvz] = \mathbb E[\vepsilon - \rvz_0\mid \rvz_t= \rvz]$.
Latent diffusion models $V_{\vtheta}$ are trained to approximate the velocity field by minimizing the flow matching loss in the latent space, 
    $\mathcal L_{\mathrm{Diff}}(\vphi^{-}, \vtheta) = \mathbb E_{\rvz_0, \vepsilon, t} \left[\|V_\vtheta(\rvz_t, t) - (\vepsilon - \rvz_0)\|_2^2\right]$,
where $\vphi^{-}$ denotes the parameter of a stop-gradient VAE encoder.

\vspace{-10pt}
\paragraph{Representation from VFMs.}
Recent studies show that incorporating semantic representations from self-supervise pretrained vision foundation models (VFMs), such as DINO~\cite{oquab2023dinov2,simeoni2025dinov3}, CLIP~\cite{radford2021clip} and MAE~\cite{he2022masked}, can substantially improve the convergence and generative quality of diffusion models. Depending on how VFMs are utilized, existing approaches fall into two main categories: representation alignment and direct representation utilization. The representation alignment family~\cite{yu2025repa,leng2025repae,lightdit} distills VFM semantics by enforcing consistency between latent (or intermediate diffusion features) and VFM representations. For instance, REPA~\cite{yu2025repa} employs a cosine-similarity objective, whereas VA-VAE~\cite{lightdit} adopts a more sophisticated similarity loss. REPA-E~\cite{leng2025repae} backpropagates a representation alignment loss through from diffusion to VAE encoder. In contrast, another line of work~\cite{rae,svg,bi2025vision,chen2025aligning} replaces the VAE encoder with a pretrained VFM and directly leverages its representation as the latent, with different design choices such as freezing the tokenizer in RAE~\cite{rae}, adding a learnable residual module in SVG~\cite{svg}, or inserting a learnable projection layer in VFM-VAE~\cite{bi2025vision}. To better integrate VFMs into our end-to-end training paradigm as a flexible plug-and-play module, we adopt the representation alignment strategy and introduce \emph{REPA++}, an enhanced alignment mechanism in our end-to-end training framework.

\section{Our Approach}\label{sec:method}
We now outline the key components of our approach. We first analyze the vanilla end-to-end training (\cref{sec:vanilla-e2e}) and identify latent collapse as the primary cause of failure. To address this issue, we introduce a simple yet effective \emph{Tweedie Pixel Reconstruction} loss that provides additional pixel-level feedback (\cref{sec:tpr-loss}). We then enhance the framework with refined architectures and REPA++ (\cref{sec:joint-scaling}, \cref{sec:repa++}) to accelerate convergence and improve final performance.

\subsection{Vanilla End-to-end Training}\label{sec:vanilla-e2e}
The vanilla end-to-end setup directly couples the VAE and diffusion model by optimizing both objectives simultaneously from scratch:
\vspace{-3pt}
\begin{equation}\label{eq:vanilla-e2e}
\mathcal L_{\text{Vanilla-E2E}}(\vphi, \vpsi, \vtheta)
= \mathcal L_{\text{VAE}}(\vphi, \vpsi) + \mathcal L_{\text{Diff}}(\vphi, \vtheta),
\end{equation}
where $\vphi$, $\vpsi$ and $\vtheta$ refers to the parameters of VAE encoder, VAE decoder and diffusion, respectively.
Despite its conceptually simple formulation, backpropagating the diffusion loss into the VAE encoder causes severe degradation of the latent space, consistent with observations in prior work~\citep{leng2025repae}.
In our experiments, the vanilla end-to-end setup attains an FID of 33.95 on ImageNet 256$\times$256 after 80 epochs, which is considerably worse than standard two-stage LDM baselines. We refer this setting as \emph{Vanilla E2E}.

\vspace{-10pt}

\paragraph{Why does Vanilla E2E training fail?}
We observe that vanilla end-to-end training drives the latent space into a highly degenerate representation. To diagnose this phenomenon, we compute the Tweedie estimate of the clean latent $\hat\rvz_0 = \rvz_t - t V_\vtheta(\rvz_t,t)$ and compare the latent reconstruction error $\|\hat\rvz_0 - \rvz_0\|^2$ with the corresponding pixel reconstruction error $\|\mathcal D_{\vpsi}(\hat \rvz_0) - \rvx_0\|^2$. As shown in \cref{fig:analysis_latent}a, the latent error is nearly zero, whereas the pixel error remains substantial. This indicates that the diffusion model fits its latent-space objective very well, yet the resulting latent code no longer preserves the information required for accurate pixel reconstruction.

To further scrutinize this failure mode, we visualize the latent distribution and find that it deviates significantly from a standard Gaussian prior, forming multiple sharp peaks in latent space (see \cref{fig:analysis_latent}b). A closer inspection reveals strong per-channel biases and heavily suppressed per-channel variances (see \cref{fig:analysis_latent}c). In such a collapsed regime, most latent codes concentrate in a few narrow regions, and the conditional distribution of clean latent $\rvz_0$ given noisy state $\rvz_t$ becomes extremely concentrated. The diffusion model can then minimize its loss by predicting nearly constant mean values across inputs, because small perturbations around these peaks incur very little error under the diffusion objective.

Taken together, these observations suggest that the primary cause of failure is an exploitable latent space: without any pixel-space feedback or explicit regularization, the diffusion model can converge to trivial solutions that achieve low diffusion loss in latent space while discarding the semantic content needed for faithful image reconstruction.

\subsection{Tweedie Pixel Reconstruction Loss}\label{sec:tpr-loss}
The latent collapse observed in vanilla end-to-end training suggests that the diffusion model can exploit the latent space for trivial solution when no pixel-level constraints are provided. To counter this, we strengthen the connection between diffusion predictions and the underlying image content so that the VAE is compelled to maintain a semantically meaningful latent representation. This motivates the \emph{Tweedie Pixel Reconstruction} (TPR) loss, which decodes the Tweedie-estimated latent back into pixel space and directly enforces consistency with the clean image.

Given the stochastic interpolant $\rvz_t = (1-t)\rvz_0 + t\vepsilon$, the diffusion model estimates the clean latent via Tweedie’s formula $\hat\rvz_0 = \rvz_t - tV_\vtheta(\rvz_t,t)$. We decode this estimate using $D_\vpsi$ and penalize discrepancies with the ground-truth image:
\begin{equation}\label{eq:tpr-loss}
\mathcal L_{\text{TPR}}(\vphi, \vpsi, \vtheta)
= \mathbb E_{\rvx_0, \rvz_0, \vepsilon, t}
\left[\left\|\mathcal D_{\vpsi}(\hat \rvz_0) - \rvx_0\right\|^2\right].
\end{equation}
This objective amounts to a pixel-space MSE and can be further enhanced with perceptual losses such as LPIPS~\citep{lpips}.

We augment the vanilla end-to-end objective in \cref{eq:vanilla-e2e} with the TPR loss. This addition effectively prevents latent collapse (see \cref{fig:analysis_latent}) and substantially improves generation quality, reducing the FID on ImageNet 256$\times$256 generation from 33.95 to 5.79 after 80 epochs. We refer this setting as \emph{E2E w/ TPR Loss}.

\subsection{Scaling with Refined Architecture}\label{sec:joint-scaling}
Building on our end-to-end framework, we adopt a vision-transformer-based VAE ({\ourvae}) in place of the conventional convolutional design (CNN-VAE). The encoder employs a patch-embedding layer followed by transformer blocks, and the decoder mirrors this structure.

For the diffusion backbone, we refine DiT~\cite{peebles2023scalable} with several modern architectural components. Following LighteningDiT~\citep{lightdit}, we integrate RMSNorm, SwiGLU activations, and 2D RoPE, and we replace block-specific modulation layers with a shared global modulation mechanism supplemented by per-block learnable biases, similar to PixArt-$\alpha$~\citep{chen2023pixartalpha}. Unless otherwise specified, the patch size is fixed to 1 in all experiments. We refer this as {\ourdit}. Additional architectural details are provided in \cref{app:model-detail}.

Employing {\ourvae} and {\ourdit} and training them end-to-end with the TPR loss, this variant reduces computation while improving performance, enabling improved scaling behavior. Model configurations are summarized in \cref{tab:model_config} in Appendix. Each training step consists of two decoder passes (for the VAE and the TPR objective) and one encoder and diffusion pass. With these architectural refinements, the total training cost decreases from 436.29 to 334.98 GFLOPs, and the model achieves an improved FID of 3.66 on ImageNet 256$\times$256 after 80 epochs. We refer to this setting as \emph{{\ours} (w/o REPA++)} to emphasize that it does not require alignment from pretrained VFMs.

\subsection{Representation Alignment via REPA++}\label{sec:repa++}
We introduce REPA++ to further enhance semantic consistency and accelerate convergence by aligning both the latent representation and the intermediate diffusion features to a pretrained VFM. Let $\rvy=\text{VFM}(\rvx_0)$ denote the semantic representation of the clean image. The first component, Latent-REPA, aligns the clean latent $\rvz_0$ with the VFM representation by mapping $\rvz_0$ through a two-layer MLP $h_{\vomega_1}$ and maximizing patch-wise cosine similarity, where patches are indexed by $k$:
\vspace{-3pt}
{
\begin{equation}\label{latent-repa}
    \mathcal L_{\text{Latent-REPA}}(\vphi,\vomega_1)
    = - \mathbb{E}_{\rvz_0, \rvx_0, k}
    \big[\text{sim}(h_{\vomega_1}^k(\rvz_0),\, \rvy^{k})\big].
\end{equation}
}
The second component, Diff-REPA, follows the formulation of \cite{leng2025repae,yu2025repa} and injects semantic guidance directly into the denoising process by aligning the intermediate diffusion feature $\rvf_t$ (extracted from an early diffusion layer given random noisy latent state $\rvz_t$) to the same VFM representation using another MLP $h_{\vomega_2}$:
{
\begin{equation}\label{diff-repa}
    \mathcal L_{\text{Diff-REPA}}(\vphi,\vtheta,\vomega_2)
    = - \mathbb{E}_{\rvz_0, \rvx_0, k}
    \big[\text{sim}(h_{\vomega_2}^k(\rvf_t),\, \rvy^k)\big].
\end{equation}
}
Finally, REPA++ combines both alignment signals through the simple addition of \cref{latent-repa} and \cref{diff-repa} with learnable parameters $\vomega=(\vomega_1,\vomega_2)$, enabling semantic supervision to act simultaneously on the latent representation and the diffusion denoising process, and consequently guiding the VAE encoder through end-to-end gradient feedback.

To this end, we have discussed all the ingradients of {\ours}. The complete training pipeline of {\ours} is shown in~\cref{fig:pipeline}, and the overall training objective is:
{\small
\begin{equation}\label{eq:seediff-loss}
\boxed{
\begin{aligned}
\mathcal L_{\text{{\ours}}}(\vphi, \vpsi,& \vtheta, \vomega)
= \mathcal L_{\text{VAE}}(\vphi, \vpsi) 
  + \mathcal L_{\text{Diff}}(\vphi, \vtheta) \\
&\quad + \mathcal L_{\text{TPR}}(\vphi, \vpsi, \vtheta)
  + \mathcal L_{\text{REPA++}}(\vphi, \vtheta, \vomega)
\end{aligned}
}
\end{equation}
}
We will refer this setting as \emph{{\ours} (w/ REPA++)} to hightlight requirement of a pretrained VFM, which achieves an further improved FID of 1.69 on ImageNet 256$\times$256 after 80 epochs.

\begin{table}[t]
\centering
\renewcommand{\arraystretch}{1.12}
\setlength{\tabcolsep}{4pt}
\caption{\textbf{From baseline to {\ours}.} We evaluate FID without guidance on ImageNet 256$\times$256 generation after 80 training epochs.}
\label{tab:comparison}
\resizebox{0.48\textwidth}{!}{%
\begin{tabular}{c!{\vline}l!{\vline}c}
\toprule
\multirow{1}{*}{\textbf{Diffusion Loss for VAE}} & 
\multirow{1}{*}{\textbf{Configuration}} & 
\textbf{gFID$\downarrow$} \\
\midrule\midrule
\multirow{4}{*}{\textbf{Detached}} 
& Two-stage Baseline & 13.02 \\
\multirow{4}{*}{{\small (w/o direct gradient feedback)}} & Single-stage Baseline & 14.21 \\
& + TPR Loss & 11.65 \\
& + Refined Architecture & 7.52 \\
& + REPA++ & 3.47 \\
\midrule\midrule
\multirow{3}{*}{\textbf{End-to-end}} 
& Vanilla E2E & 33.95 \\
\multirow{3}{*}{{\small (w/ direct gradient feedback)}} & + TPR Loss & 5.79 \\
& + Refined Architecture & 3.66 \\
& + REPA++  & \textbf{1.69} \\
\bottomrule
\end{tabular}
}
\vspace{-8pt}
\end{table}

\definecolor{rowhl}{HTML}{E6E6FA}

\begin{table*}[t]
\centering
\renewcommand{\arraystretch}{1.12}
\setlength{\tabcolsep}{4pt}

\caption{\textbf{ImageNet 256$\times$256 generation benchmark.} 
We report detailed results for both non–representation alignment and representation alignment methods. We additionally include results at 80 epochs for convergence speed comparison. {\ours} achieves state-of-the-art performance without guidance in both settings ($\downarrow$ lower is better; $\uparrow$ higher is better).}

\label{tab:imagenet-256}
\resizebox{\textwidth}{!}{%
\begin{tabular}{l!{\vline}l!{\vline}l!{\vline}l!{\vline}l!{\vline}c!{\vline}*{5}{c}!{\vline}*{5}{c}c}
\toprule
\multirow{2}{*}{\textbf{Method}} &
\multicolumn{2}{c!{\vline}}{\textbf{VAE}} &
\multicolumn{2}{c!{\vline}}{\textbf{Diffusion}} &
\multirow{2}{*}{\textbf{Epochs}} &
\multicolumn{5}{c!{\vline}}{\textbf{Generation w/o guidance}} &
\multicolumn{5}{c}{\textbf{Generation w/ guidance}} &
\multirow{2}{*}{\textbf{rFID$\downarrow$}} \\
\cmidrule(lr){2-3}\cmidrule(lr){4-5}\cmidrule(lr){7-11}\cmidrule(lr){12-16}
& \textbf{Model} & \textbf{Params (M)} & \textbf{Model} & \textbf{Params (M)} & &
\textbf{gFID$\downarrow$} & \textbf{sFID$\downarrow$} & \textbf{IS$\uparrow$} & \textbf{Prec.$\uparrow$} & \textbf{Rec.$\uparrow$} &
\textbf{gFID$\downarrow$} & \textbf{sFID$\downarrow$} & \textbf{IS$\uparrow$} & \textbf{Prec.$\uparrow$} & \textbf{Rec.$\uparrow$} & \\
\midrule
\midrule
\rowcolor{rowhl}
\multicolumn{17}{c}{\textbf{Non-representation Alignment Methods}} \\
\midrule
\midrule
MaskDiT~\cite{maskdit} & \multirow{6}{*}{SD-VAE} & \multirow{6}{*}{84} & \multirow{6}{*}{DiT-XL/2} &\multirow{6}{*}{675} & 1600 & 5.69 & 10.34 & 177.9 & 0.74 & 0.60 & 2.28 & 5.67 & 276.6 & 0.80 & 0.61 & \multirow{6}{*}{0.61} \\
DiT~\cite{peebles2023scalable} &  &  &  &  & 1400 & 9.62 & 6.85  & 121.5 & 0.67 & 0.67 & 2.27 & 4.60 & 278.2 & \hldigit{0.83} & 0.57 &  \\
SiT~\cite{ma2024sit} &  &  &  &  & 1400 & 8.61 & 6.32  & 131.7 & 0.68 & 0.67 & 2.06 & 4.50 & 270.3 & 0.82 & 0.59 &  \\
FasterDiT~\cite{yao2024fasterdit} &  &  &  &  & 400 & 7.91 & 5.45  & 131.3 & 0.67 & \hldigit{0.69} & 2.03 & 4.63 & 264.0 & 0.81 & 0.60 &  \\
MDT~\cite{gao2023masked} &  &  &  &  & 1300 & 6.23 & 5.23  & 143.0 & 0.71 & 0.65 & 1.79 & 4.57 & 283.0 & 0.81 & 0.61 &  \\
MDTv2~\cite{gao2023mdtv2} &  &  &  &  & 1080 & \NA  & \NA   & \NA   & \NA  & \NA  & 1.58 & 4.52 & 314.7 & 0.79 & \hldigit{0.65} &  \\
\midrule
\midrule

\multirow{6}{*}{{\ours}}  & \multirow{2}{*}{{{\ourvae}-B}} & \multirow{2}{*}{58} & \multirow{2}{*}{{{\ourdit}-B}} & \multirow{2}{*}{91} & 80 & 10.65   & 5.67   & 105.59  & 0.71  & 0.59 & 3.86 & 5.13 & 198.07 & 0.81 & 0.54 & 0.74 \\
\multirow{6}{*}{(w/o REPA++)} &  &  &  &  & 200 & 7.27   & 5.39   & 129.38  & 0.72  & 0.61 & 2.64 & 4.90 & 236.39 & 0.81 & 0.56 & 0.62 \\
\cmidrule(lr){2-17}
& \multirow{2}{*}{{{\ourvae}-L}} & \multirow{2}{*}{207} & \multirow{2}{*}{{{\ourdit}-L}} & \multirow{2}{*}{319} & 80 & 5.31   & 4.78   & 144.62  & 0.76  & 0.61 & 2.15 & 4.65 & 245.47 & 0.82 & 0.58 & 0.58 \\
&  &  &  &  & 200 & 3.35   & 4.62   & 176.40  & 0.78  & 0.63 & 1.61 & 4.51 & 281.22 & 0.81 & 0.61 & 0.50 \\
\cmidrule(lr){2-17}
& \multirow{2}{*}{{{\ourvae}-XL}} & \multirow{2}{*}{437} & \multirow{2}{*}{{{\ourdit}-XL}} & \multirow{2}{*}{672} & 80 & 3.66  & 4.42 & 172.03    & 0.78  & 0.62 & 1.80 & 4.40 & 267.10 & 0.81 & 0.59 & 0.54 \\
&  &  &  &  & 200 & \hldigit{2.42}  & \hldigit{4.38} & \hldigit{197.84}  & \hldigit{0.80}  & 0.63 & \hldigit{1.46} & \hldigit{4.32} &\hldigit{ 297.13} & 0.81 & 0.62 & \hldigit{0.49} \\

\midrule
\midrule
\rowcolor{rowhl}
\multicolumn{17}{c}{\textbf{Representation Alignment Methods}} \\
\midrule
\midrule
 \multirow{2}{*}{REPA~\cite{yu2025repa}} & \multirow{2}{*}{SD-VAE} & \multirow{2}{*}{84} & \multirow{2}{*}{DiT-XL/2} & \multirow{2}{*}{675} &  80 & 7.90 & 5.06 & 122.6 & 0.70 & 0.65 & \NA  & \NA  & \NA  & \NA & \NA & \multirow{2}{*}{0.61}\\
 &  &  &  &  & 800 & 5.90 & 5.73 & 157.8 & 0.70 & \hldigit{0.69} & 1.42 & 4.70 & 305.7 & 0.80 & 0.65 &  \\
\midrule
 \multirow{2}{*}{REG~\cite{wu2025representation}} & \multirow{2}{*}{SD-VAE} & \multirow{2}{*}{84} & \multirow{2}{*}{DiT-XL/2} & \multirow{2}{*}{675} &  80 & 3.40 & \NA & \NA & \NA & \NA & 1.86 & 4.49 & 321.4 & 0.76 & 0.63 & \multirow{2}{*}{\hldigit{0.28}}\\
 &  &  &  &  & 480 & 2.20 & \NA & \NA & \NA & \NA & 1.40 & 4.26 & 296.9 & 0.77 & \hldigit{0.66} &  \\
 \midrule
 \multirow{2}{*}{REPA-E~\cite{leng2025repae}} & \multirow{2}{*}{E2E-VAE} & \multirow{2}{*}{70} & \multirow{2}{*}{DiT-XL/1} & \multirow{2}{*}{675} &  80 & 3.46 & 4.17 & 159.8 & 0.77 & 0.63 & 1.67 & 4.12 & 266.3 & 0.80 & 0.63 & \multirow{2}{*}{\hldigit{0.28}}\\
 &  &  &  &  & 800 & 1.83 & 4.22 & 217.3 & 0.77 & 0.66 & 1.26 & 4.11 & 314.9 & 0.79 & \hldigit{0.66} &  \\

\midrule
\midrule

\multirow{6}{*}{{\ours}} & \multirow{2}{*}{{{\ourvae}-B}} & \multirow{2}{*}{58} & \multirow{2}{*}{{{\ourdit}-B}} & \multirow{2}{*}{91} & 80 & 6.20 & 5.12 & 138.71  & 0.74  & 0.60 & 2.29 & 4.83 & 230.33 & 0.79 & 0.57 & 0.84 \\
\multirow{6}{*}{(w/ REPA++)}&  &  &  &  & 200 & 4.44  & 4.94  & 156.90  & 0.79  & 0/60 & 1.81 & 4.72 & 244.37 & 0.79 & 0.60 & 0.71 \\
\cmidrule(lr){2-17}
& \multirow{2}{*}{{{\ourvae}-L}} & \multirow{2}{*}{207} & \multirow{2}{*}{{{\ourdit}-L}} & \multirow{2}{*}{319} & 80 & 2.15   & 4.21   & 207.87  & 0.79  & 0.61 & 1.41 & 4.15 & 292.90 & 0.80 & 0.61 & 0.62 \\
&  &  &  &  & 200 & 1.65  & 4.20   & 232.74  & 0.80  & 0.63 & 1.25 & 4.13 & 315.55 & 0.80 & 0.62 & 0.55 \\
\cmidrule(lr){2-17}
& \multirow{2}{*}{{{\ourvae}-XL}} & \multirow{2}{*}{437} & \multirow{2}{*}{{{\ourdit}-XL}} & \multirow{2}{*}{672} & 80 & 1.69   & \hldigit{4.14}   & 234.56  & 0.80  & 0.63 & 1.29 & \hldigit{4.08} & 309.42 & 0.80 & 0.63 & 0.59 \\
&  &  &  &  & 200 & \hldigit{1.50}   & 4.18   & 253.09  & \hldigit{0.81}  & 0.64 & \hldigit{1.21} & 4.11 & \hldigit{317.09} & \hldigit{0.80} & 0.63 & 0.52 \\

\bottomrule
\end{tabular}
}%
\end{table*}

\definecolor{rowhl}{HTML}{E8F3FF}

\begin{table*}[t]
\centering
\renewcommand{\arraystretch}{1.12}
\setlength{\tabcolsep}{4pt}
\caption{\textbf{ImageNet 512$\times$512 generation benchmark.} We employ a more efficient 32$\times$ downsampled VAE while keeping all other settings consistent with the 256$\times$256 setup. {\ours} offers flexible VAE configurations and enables efficient high-resolution generation without sacrificing performance ($\downarrow$ lower is better; $\uparrow$ higher is better).}

\label{tab:imagenet-512}
\resizebox{\textwidth}{!}{%
\begin{tabular}{l!{\vline}l!{\vline}l!{\vline}l!{\vline}l!{\vline}l!{\vline}l!{\vline}c!{\vline}*{3}{c}!{\vline}*{3}{c}}
\toprule
\multirow{2}{*}{\textbf{Method}} &
\multicolumn{3}{c!{\vline}}{\textbf{VAE}} &
\multicolumn{3}{c!{\vline}}{\textbf{Diffusion}} &
\multirow{2}{*}{\textbf{Epochs}} &
\multicolumn{3}{c!{\vline}}{\textbf{Generation w/o guidance}} &
\multicolumn{3}{c}{\textbf{Generation w/ guidance}}  \\
\cmidrule(lr){2-4}\cmidrule(lr){5-7}\cmidrule(lr){9-11}\cmidrule(lr){12-14}
& \textbf{Model} & \textbf{Params (M)} & \textbf{FLOPs (G)} & \textbf{Model} & \textbf{Params (M)} & \textbf{FLOPs (G)} & &
\textbf{gFID$\downarrow$} & \textbf{sFID$\downarrow$} & \textbf{IS$\uparrow$} & 
\textbf{gFID$\downarrow$} & \textbf{sFID$\downarrow$} & \textbf{IS$\uparrow$} \\
\midrule
\midrule
U-ViT~\cite{bao2023all} & \multirow{6}{*}{SD-VAE} & \multirow{6}{*}{84} & \multirow{6}{*}{541 + 1240} & U-ViT-H/4 & 501 & 133 & 400 & - & - & - & 4.05 & 6.44 & 263.79\\
DiT~\cite{peebles2023scalable} &  &  &  & DiT-XL/2 & 675 & 524 & 600 & 12.03 & 7.12  & 105.25  & 3.04  & 5.02  & 240.82\\
SiT~\cite{ma2024sit} &  &  &  & DiT-XL/2 & 675 & 524 & 600 & - & -  & -  & 2.62  & 4.18  & 252.21 \\
MaskDiT~\cite{maskdit} &  &  &  & DiT-XL/2 & 675 & 524 & 800 & 10.79 & 13.41  & 145.08  & 2.50  & 5.10  & 256.27\\
EDM2~\cite{karras2024analyzing} &  &  &  & EDM2-XS & 125 & 46 & 1600 & 3.53 & -  & -  & 2.91  & -  & - \\
EDM2\cite{karras2024analyzing} &  &  &  & EDM2-XXL & 1523 & 406 & 1600 & 1.91 & -  & -  & 1.81  & -  & - \\
\midrule\midrule

\multirow{2}{*}{{\ours}}  & {\ourvae}-B & 61 & 8+8 & {\ourdit}-B & 91 & 22 & 200 & 7.43 & 5.96  & 131.42  & 2.98  & 5.42  & 180.92 \\
\multirow{2}{*}{(w/o REPA++)}  & {\ourvae}-L & 212 & 27+27 & {\ourdit}-L & 319 & 79 & 200 & 2.97 & 4.81  & 181.58  & 1.87  & 4.46  & 223.57 \\
  & {\ourvae}-XL & 444 & 57+57 & {\ourdit}-XL & 672 & 167 & 200 & 2.18 & 4.23  & 218.26  & 1.79 & 4.18  & 231.89 \\

\midrule

\multirow{2}{*}{{\ours}}  & {\ourvae}-B & 61 & 8+8 & {\ourdit}-B & 91 & 22 & 200 & 4.89 & 5.56  & 162.69  & 2.19  & 5.02  & 213.66 \\
\multirow{2}{*}{(w/ REPA++)}  & {\ourvae}-L & 212 & 27+27 & {\ourdit}-L & 319 & 79 & 200 & 1.92 & 4.27  & 210.04  & 1.34  & 4.22  & 309.57 \\
  & {\ourvae}-XL & 444 & 57+57 & {\ourdit}-XL & 672 & 167 & 200 & \hldigit{1.53} & \hldigit{4.07}  & \hldigit{258.47}  & \hldigit{1.26}  & \hldigit{4.03}  & \hldigit{322.55} \\

\bottomrule
\end{tabular}
}%
\vspace{-2pt}
\end{table*}

\section{Experiments}
We first outline the experimental setup in \cref{sec:setup}, followed by a controlled component-wise comparison in \cref{sec:component-wise-comparison}. We then systematically examine the impact of end-to-end training on both the diffusion model (\cref{sec:impact-for-diffusion}) and the VAE latent space (\cref{sec:impact-for-latent}). Key ablations are further summarized in \cref{sec:ablations}. More results are shown in \cref{app:add-results}.

\begin{figure*}[!t]
    \centering
    \includegraphics[width=\linewidth]{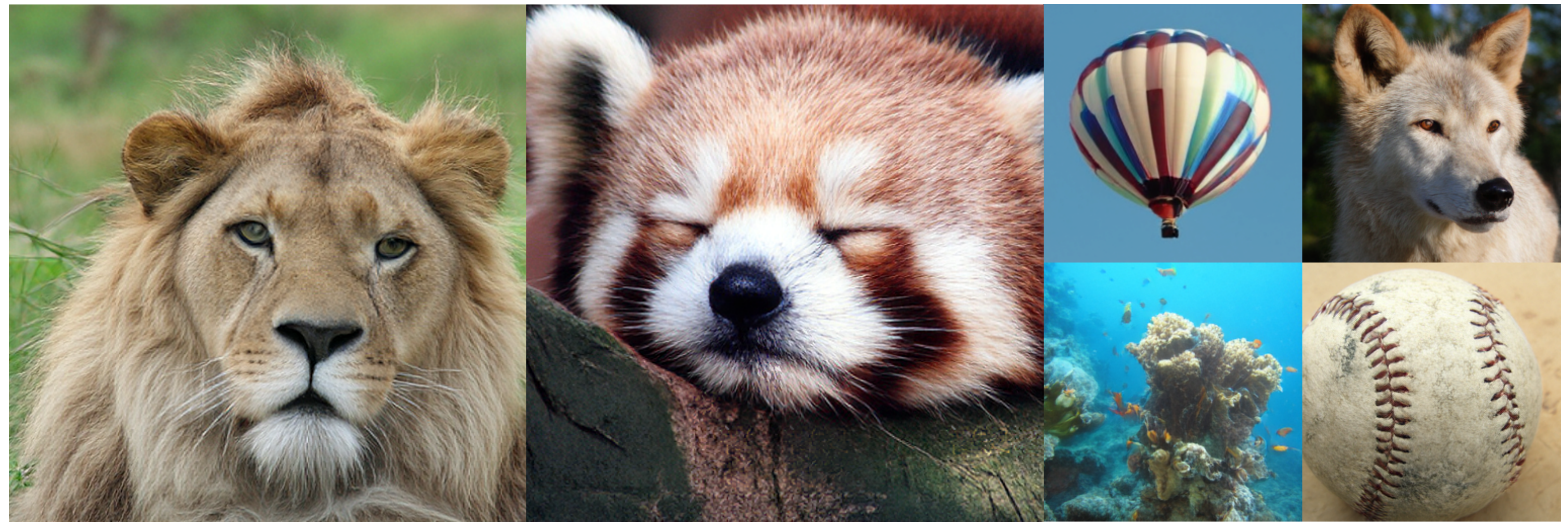}
    \caption{\textbf{Selected class-conditioned samples} from {\ours}-XL at 512$\times$512 (left) and 256$\times$256 (right) resolutions without guidance.
}
    \label{fig:main_figure}
\end{figure*}

\subsection{Setup}\label{sec:setup}

\paragraph{Training.}
We follow the training setup of SiT~\cite{ma2024sit} and REPA~\cite{yu2025repa} for ImageNet~\cite{deng2009imagenet} 256$\times$256 generation. The dataset is preprocessed with center cropping~\cite{he2016deep} and random horizontal flipping~\cite{krizhevsky2012imagenet}. All models are trained for 200 epochs (1M iterations) with a batch size of 256, which we find sufficient for {\ours} to converge.  
We use the Adam optimizer~\cite{kingma2014adam} with a constant learning rate of 1e{-4} across all components. Exponential moving averages (EMA)~\cite{polyak1992acceleration} are maintained for {\ourvae} and {\ourdit} with decay rates of 0.999 and 0.9999, respectively.  
We consider three model scales (B, L, XL) for both {\ourvae} and {\ourdit}, with parameters, FLOPs, and memory consumption summarized in \cref{tab:model_config} in Appendix. All {\ourvae} models adopt a 16$\times$ spatial compression and a 32 latent channels (f16d32). For 512$\times$512 generation, we modify only the first patch embedding layer to use 32$\times$ spatial compression (f32d32), keeping all other configurations unchanged.  
We use DINOv3-ViT-L/16 (0.3B) as the default VFM for representation alignment. To ensure reproducibility, we include the training pseudocode in~\cref{alg:main} in Appendix, and list detailed model configurations and hyperparameters in~\cref{tab:detailed_config} in Appendix. Implementation details of all loss objectives are provided in \cref{app:detailed-imple}.

\vspace{-10pt}
\paragraph{Sampling.}
For class-conditioned generation, we adopt the standard Euler–Maruyama SDE solver~\cite{sysde} with 250 NFEs by default. Guidance sampling uses AutoGuidance (AG)~\cite{auto-guidance} instead of Classifier-Free Guidance (CFG)~\cite{cfg}, as we observe that the benefit of CFG diminishes when the diffusion model attains strong unconditional performance with the null-class embedding. AG addresses this issue by leveraging a weaker auxiliary model that does not rely on null-class conditioning. Further implementation details and rationale are discussed in \cref{app:guidance}.

\vspace{-10pt}
\paragraph{Evaluation.}
For image generation, following common practice, we report generation Frechet Inception Distance (gFID)~\cite{heusel2017gans}), spatial FID (sFID)~\cite{nash2021generating}, Inception Score (IS)~\cite{salimans2016improved}), precision (Pre.) and recall (Rec.)~\cite{kynkaanniemi2019improved}, all computed with 50{,}000 generated samples. Unless specified, all images are generated without guidance. We perform evaluation using the official OpenAI diffusion model evaluation suite~\cite{dhariwal2021diffusion}. To assess the reconstruction capability of the VAE, we additionally report the reconstruction FID (rFID) and Peak Signal-to-Noise Ratio (PSNR)~\cite{wang2006modern}. For evaluating semantics of the learned latent space and its alignment to pretrained VFM representations, we further follow REPA~\cite{yu2025repa} and report linear probing accuracy and the CKNNA score~\cite{huh2024platonic}.

\begin{figure*}[!t]
    \centering
    \includegraphics[width=\linewidth]{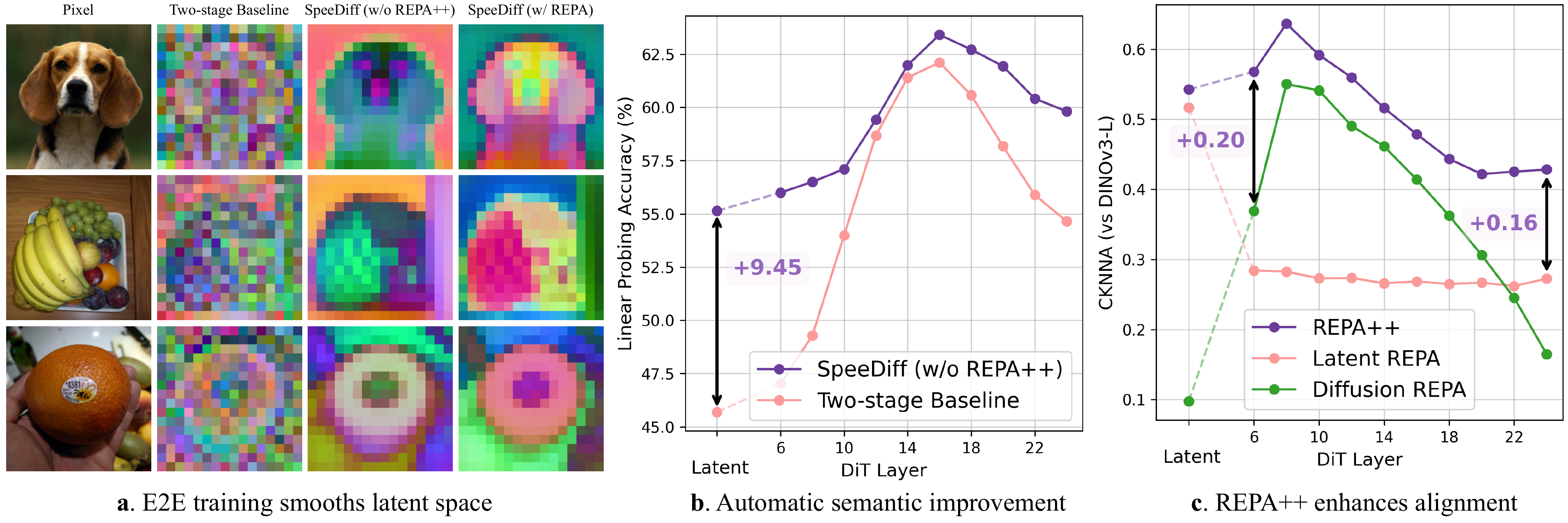}
    \caption{\textbf{Analysis of latent and intermediate diffusion features.} (a) Visualization of latents using Robust PCA \cite{candes2011robust} on 1000 images. (b) Linear probing accuracy on the ImageNet validation set, which shows end-to-end training with diffusion gradient feedback automatically improve semantics. (c) CKNNA evaluation of latent and intermediate diffusion features for the REPA++ components.}
    \label{fig:semantics}
    \vspace{-10pt}
\end{figure*}

\subsection{Component-wise Comparison}\label{sec:component-wise-comparison}
We begin by examining the contribution of each component and, in particular, the effect of allowing diffusion loss gradients to update the VAE encoder. Beyond the configurations described in \cref{sec:method}, we introduce more baselines to provide clearer reference points.  
The first is \emph{Two-stage Baseline} of LDM, where a convolutional VAE (CNN-VAE) is trained to convergence before training a DiT-XL/1 diffusion model on its latent space. To isolate the effect of end-to-end gradient feedback, we also include a \emph{One-stage Baseline} in which the VAE and diffusion model are trained jointly from scratch, but a stop-gradient operation prevents diffusion loss from propagating into the VAE encoder.  
In summary, as shown in \cref{tab:comparison}, once TPR loss is applied to prevent latent collapse, end-to-end optimization consistently outperforms both detached variants.

\subsection{Impact of E2E Training for Diffusion}\label{sec:impact-for-diffusion}
\paragraph{Accelerated convergence.}
A primary advantage of {\ours} lies in its ability to substantially accelerate diffusion model convergence. On ImageNet 256$\times$256, {\ours}-XL (w/ REPA++) reaches a gFID of 7.36 without guidance within 10 epochs, already surpassing the vanilla SiT baseline trained for 1400 epochs.  
As shown in \cref{tab:imagenet-256}, {\ours}-XL (w/o REPA++) attains a gFID of 3.66 without guidance after 80 epochs, indicating that allowing diffusion loss gradients to directly shape the latent space significantly improves convergence speed.  
The incorporation of REPA++ further strengthens this effect, enabling {\ours}-XL to reach a gFID of 1.69 without guidance after 80 epochs. {\ours} could produce high visual fidelity without guidance, with representative samples presented in \cref{fig:main_figure}.

\vspace{-10pt}
\paragraph{Improved scaling behavior.}
{\ours} also induces improved scaling behavior in diffusion models. As shown in \cref{fig:teaser} and \cref{tab:imagenet-256,tab:imagenet-512}, {\ours} exhibits a clearer and more monotonic scaling trend compared to prior state of the art approaches such as EDM2~\cite{karras2024analyzing}. {\ours} adopts a fully transformer based pipeline, which benefits from the strong scaling properties and avoids the capacity limitations of convolution. In addition, {\ours} scales the VAE and diffusion model jointly rather than enlarging only the diffusion while keeping the VAE fixed. This joint scaling leads to improved scaling behavior. Further analysis in \cref{app:joint-scaling} examines parameter allocation across the VAE encoder, VAE decoder, and diffusion model, and shows that appreciate allocation yields consistently stronger performance.

\subsection{Impact of E2E Training for VAE Latent}\label{sec:impact-for-latent}

\paragraph{Automatic semantic gain.}
Analysis of the latent spaces of {\ours} (w/o REPA++) and {\ours} (w/ REPA++) shows that end-to-end training with diffusion loss gradients naturally smooths the latent and strengthens its semantic organization, as illustrated in \cref{fig:semantics}a, where independently trained VAEs tend to overfit low level image statistics and exhibit weak semantic separation. This phenomenon is further supported by the quantitative results in \cref{fig:semantics}b, where latent linear probing accuracy on ImageNet improves by 9.45\%, indicating that the diffusion objective induces a more semantically aligned latent space.

\vspace{-10pt}
\paragraph{Generation-friendly latent space.}
We further examine whether the trained VAE in {\ours} remains effective when used separately from the diffusion model it was trained with. We freeze the converged VAE from {\ours}-XL (w/ REPA++) and train a new diffusion model from scratch, which converges at a similar speed to {\ours} (\cref{tab:frozen-vae}). This suggests that {\ours} learns a generation-friendly latent space. From this perspective, {\ours} can be interpreted as performing latent pretraining, producing representations inherently better suited for downstream generative modeling.

\begin{table}[t]
\centering
\caption{\textbf{Can {\ours} pretrained VAE be used independently?} A diffusion model trained on top of the frozen {\ours} pretrained VAE converges at nearly the same rate as {\ours} itself. Results are reported on ImageNet 256$\times$256.}
\label{tab:frozen-vae}
\renewcommand{\arraystretch}{1.12}
\setlength{\tabcolsep}{4pt}
\resizebox{0.48\textwidth}{!}{%
    \begin{tabular}{lcccc}
    \toprule
      \textbf{Configuration} & \textbf{gFID$\downarrow$} (10 epochs) & \textbf{gFID$\downarrow$} (80 epochs) \\
      \midrule
     {\ours}-XL Pretrained VAE & 6.94 & 1.73\\
      \textit{Reference}: {\ours}-XL & 7.36 & 1.69 \\
      \bottomrule
    \end{tabular}
} %
\end{table}

\begin{table}[t]
\centering
  \setlength{\tabcolsep}{4pt}
   \caption{\textbf{Ablation on REPA++.} We evaluate different variants of REPA++ components on ImageNet 256$\times$256. REPA++ with DINOv3-L yields the best overall performance.}
    \label{tab:repa-ablation}
\resizebox{0.42\textwidth}{!}{%
    \begin{tabular}{lc}
    \toprule
      \textbf{Configuration} & \textbf{gFID$\downarrow$} (80 epochs) \\
      \midrule
      REPA++ (DINOv3-ViT-L/16~\cite{simeoni2025dinov3}) & \textbf{1.69} \\
       \midrule
      REPA++ (DINOv2-ViT-L/14~\cite{oquab2023dinov2}) & 1.76 \\
      Latent-REPA (DINOv3-ViT-L/16~\cite{simeoni2025dinov3}) & 2.28 \\
      Diff-REPA (DINOv3-ViT-L/16~\cite{simeoni2025dinov3}) & 2.15 \\
      \bottomrule
    \end{tabular}
}
\vspace{-5pt}
\end{table}

\begin{table}[t]
\centering
\caption{\textbf{Ablations on latent channels.} Results show that 32 channels offer the best trade-off between reconstruction and generation. The results are evaluate on ImageNet 256$\times$256 after 80 epochs.}
\label{tab:latent-channels}
\renewcommand{\arraystretch}{1.12}
\setlength{\tabcolsep}{4pt}
\resizebox{0.48\textwidth}{!}{%
\begin{tabular}{c!{\vline}c!{\vline}c!{\vline}cccc}
\toprule
\multicolumn{2}{c!{\vline}}{\textbf{Model}} & \textbf{Metrics} & \multicolumn{4}{c!}{\textbf{Latent Channels}} \\
\cmidrule{1-2}\cmidrule{4-7}
\textbf{VAE} & \textbf{Diffusion} & (80 epochs) & 16 & 32 & 64 & 128 \\
\midrule\midrule

\multirow{3}{*}{{\ourvae}-B} & \multirow{3}{*}{{\ourdit}-B} & \textbf{gFID$\downarrow$} & 7.43 & \textbf{6.20} & 7.66 & 12.06 \\
& & \textbf{rFID$\downarrow$} & 1.36 & 0.84 & 0.42 & \textbf{0.26} \\
& & \textbf{PSNR$\uparrow$} & 24.01 & 25.12 & 26.36 & \textbf{27.07} \\

\midrule
\multirow{3}{*}{{\ourvae}-B} & \multirow{3}{*}{{\ourdit}-XL} & \textbf{gFID$\downarrow$} & 3.02 & \textbf{2.29} & 2.87 & 4.69 \\
& & \textbf{rFID$\downarrow$} & 1.24 & 0.78 & 0.53 & \textbf{0.28} \\
& & \textbf{PSNR$\uparrow$} & 24.03 & 25.28 & 26.33 & \textbf{27.16} \\

\midrule
\multirow{3}{*}{{\ourvae}-XL} & \multirow{3}{*}{{\ourdit}-XL} & \textbf{gFID$\downarrow$} & 2.82 & \textbf{1.69} & 2.27 & 3.26 \\
& & \textbf{rFID$\downarrow$} & 0.89 & 0.59 & 0.34 & \textbf{0.20} \\
& & \textbf{PSNR$\uparrow$} & 24.67 & 25.53 & 26.44 & \textbf{27.38} \\

\bottomrule
\end{tabular}
} %
\vspace{-5pt}
\end{table}

\subsection{Ablations}\label{sec:ablations}

\paragraph{Components of REPA++.}
To understand the roles of the two components within REPA++, we evaluate Latent-REPA and Diff-REPA using the CKNNA metric computed with DINOv3-L. Latent-REPA notably improves the semantic alignment of the VAE latent space, although this benefit transfers weakly to the diffusion model. Diff-REPA, on the other hand, substantially improves the alignment of intermediate diffusion features, but the effect diminishes in deeper layers.  
REPA++ integrates these two components through a simple additive formulation, allowing their strengths to complement each other. As shown in \cref{fig:semantics}c, the latent space inherits strong alignment from Latent-REPA, while the diffusion model maintains consistently high alignment across layers due to Diff-REPA. Quantitative results are provided in \cref{tab:repa-ablation}.

\vspace{-10pt}
\paragraph{Latent channels.}
The number of latent channels serves as a key hyperparameter governing the trade-off between reconstruction and generation. Increasing the latent channels generally alleviates the information bottleneck and improves reconstruction. However, excessively large latent channels often lead to a less compact and less semantically structured latent space, making the diffusion model more difficult to train.  
Across a range of model sizes and settings, we observe that 32 channels consistently yield the best overall performance, shown in \cref{tab:latent-channels}.

\section{Conclusion}
In this work, we conduct a detailed analysis of why vanilla end-to-end training of the VAE and diffusion model fails, and identify latent space collapse as the primary cause. By introducing the Tweedie Pixel Reconstruction loss, we establish a stable and effective optimization pathway that enables the VAE to benefit directly from diffusion loss gradient supervision. Building on this foundation, architectural refinements together with the proposed REPA++ further improve convergence speed and generative quality. Overall, {\ours} sets new \emph{state-of-the-art} results for both non-alignment and VFM-aligned latent diffusion models.

\newpage
{
    \small
    \bibliographystyle{ieeenat_fullname}
    \bibliography{main}
}

\clearpage
\appendix
\setcounter{page}{1}
\onecolumn
    {\centering
    \Large
    \textbf{\thetitle}\\
    \vspace{0.5em}Supplementary Material \\
    \vspace{1.5em}
}
\section{Detailed Implementation of {\ours}}\label{app:detailed-imple}

This section provides comprehensive implementation details of {\ours} to ensure full reproducibility. We describe our model architectures in \cref{app:model-detail}, training configurations in \cref{app:training-detail}, and sampling setup in \cref{app:sampling-detail}. A complete summary of hyperparameters and model configurations is included in \cref{tab:detailed_config}.

\subsection{Model Details}\label{app:model-detail}

\subsubsection{ViT-VAE}\label{app:vit-vae}

We adopt a vision-transformer–based VAE (ViT-VAE) in place of the conventional convolutional VAE, as transformer architectures have better scalibility and offer improved compute efficiency (see \cref{tab:model_config} for comparison). 

\vspace{-10pt}
\paragraph{Encoder.}
The encoder begins with a patch-embedding layer that tokenizes each $p\times p$ patch into a hidden vector. We use $p{=}16$ for ImageNet 256$\times$256 and $p{=}32$ for 512$\times$512 generation. The embedded tokens are processed by a stack of standard transformer blocks equipped with 2D RoPE~\cite{rope-embedding} positional encoding and SwiGLU~\cite{glu} feedforward layers. A final linear projection maps the output tokens into a compact latent representation of dimension $d{=}32$.

\vspace{-10pt}
\paragraph{Decoder.}
The decoder mirrors the encoder. It first maps the latent tokens back to the hidden dimension through a linear layer, followed by transformer blocks, and finally an unpatch-embedding layer that reconstructs pixel-space outputs. To further refine visual details, we append a single convolutional layer initialized as an identity mapping. This lightweight refinement consistently improves visual quality.

\subsubsection{Refined-DiT}\label{app:refined-dit}

Our diffusion model, {\ourdit}, is based on recent architectural advances from Lightening DiT~\cite{lightdit} and incorporates a more parameter-efficient modulation mechanism. 
Due to spatial downsampling introduced by the VAE, we fix the diffusion model’s patch size to $1$, implemented as a simple linear projection into the transformer hidden dimension. The model then applies a sequence of diffusion transformer blocks. Instead of per-block modulation, we adopt a global modulation layer combined with learnable per-block biases, following the design strategy of PixArt-$\alpha$~\cite{chen2023pixartalpha}. This approach reduces parameters and computational overhead while preserving generative performance.

\begin{table*}[h!]
\centering
\renewcommand{\arraystretch}{1.12}
\setlength{\tabcolsep}{4pt}
\caption{\textbf{Model configurations.} Comparison between our refined architecture ({\ourvae}+{\ourdit}) and the baseline in terms of parameters, FLOPs, and inference memory per image. ViT-VAE is more scalable and memory friendly in our end-to-end training framework.}
\label{tab:model_config}
\resizebox{\textwidth}{!}{%
\begin{tabular}{l!{\vline}ccc!{\vline}ccc!{\vline}ccc}
\toprule
\multirow{2}{*}{\textbf{Configuration}} &
\multicolumn{3}{c}{\textbf{{\ourvae} Encoder} (ours)} &
\multicolumn{3}{c}{\textbf{{\ourvae} Decoder} (ours)} &
\multicolumn{3}{c}{\textbf{{\ourdit}} (ours)} \\
\cmidrule(lr){2-4}\cmidrule(lr){5-7}\cmidrule(lr){8-10}
& Params (M) & FLOPs (G) & Memory (M) & Params (M) & FLOPs (G) & Memory (M) & Params (M) & FLOPs (G) & Memory (M)  \\
\midrule
Base & 28.98 & 7.41 & 11.0 & 28.95 & 7.41 & 11.78 & 91.38 & 21.79 & 10.29 \\
Large & 103.68 & 26.54 & 14.75 & 103.65 & 26.54 & 20.03 & 319.38 & 79.01 & 14.06  \\
XL & 218.63 & 55.97 & 19.88 & 218.58 & 55.97 & 28.78 & 672.23 & 167.07 & 19.20  \\
\midrule
\multirow{3}{*}{\textbf{Configuration}} &
\multicolumn{3}{c}{\textbf{CNN-VAE Encoder}} &
\multicolumn{3}{c}{\textbf{CNN-VAE Decoder}} &
\multicolumn{3}{c}{\textbf{DiT-XL/1}} \\
\cmidrule(lr){2-4}\cmidrule(lr){5-7}\cmidrule(lr){8-10}
& Params (M) & FLOPs (G) & Memory (M) & Params (M) & FLOPs (G) & Memory (M) & Params (M) & FLOPs (G) & Memory (M)  \\
\midrule
- & 28.42 & 69.01 & 193.31 & 41.42 & 126.43 & 161.62 & 675.15 & 114.42 & 12.41  \\
\bottomrule
\end{tabular}
}%
\end{table*}

 \newpage
\subsection{Training Details}\label{app:training-detail}

\begin{algorithm}[htbp]
\caption{\textbf{\ours}: End-to-end pixel-anchored joint training of latent diffusion model}
\label{alg:main}
\small
\begin{algorithmic}
\Require Data $p_{\mathrm{data}}$, encoder $p_{\vphi}(\rvz_0\!\mid\!\rvx_0)$, decoder $\mathcal D_{\vpsi}$, diffusion $V_{\vtheta}$, two MLP adapters $h_{\omega_1}$ and $h_{\omega_2}$ (optional)
\While{not converged}
  \State $\rvx_0 \sim p_{\mathrm{data}},\quad \rvz_0 \sim p_{\vphi}(\rvz_0\!\mid\!\rvx_0)$
  \Statex \(\triangleright\) \textbf{VAE Reconstruction}
  \State $\hat\rvx_{\mathrm{vae}}=\mathcal D_{\vpsi}(\rvz_0),\quad L_{\mathrm{VAE}}=\mathcal L_{\mathrm{VAE}}(\hat\rvx_{\mathrm{vae}},\rvx_0;\vphi,\vpsi)$
  \State 
  \Statex \(\triangleright\) \textbf{Diffusion Flow Matching}
  \State $t\sim\mathcal U(0,1),\ \vepsilon\sim\mathcal N(\vzero,\mI)$;\ 
         $\rvz_t=(1-t)\rvz_0+t\vepsilon,\ \rvv_t=\vepsilon-\rvz_0$
  \State $\rvv_{\mathrm{pred}}=V_{\vtheta}(\rvz_t,t),\quad L_{\mathrm{Diff}}=\mathcal L_{\mathrm{Diff}}(\rvv_{\mathrm{pred}},\rvv_t;\vphi,\vtheta)$
  \State 
  \Statex \(\triangleright\) \textbf{Tweedie Pixel Reconstruction}
  \State $\rvz_{\mathrm{tweedie}}=\rvz_t-t\,\rvv_{\mathrm{pred}},\quad \hat\rvx_{\mathrm{tpr}}=\mathcal D_{\vpsi}(\rvz_{\mathrm{tweedie}})$
  \State $L_{\mathrm{TPR}}=\mathcal L_{\mathrm{TPR}}(\hat\rvx_{\mathrm{tpr}},\rvx_0; \vphi,\vpsi,\vtheta)$
  \State 

  \Statex \(\triangleright\) \textbf{REPA++ (optional)}
  \State $\rvy = \mathrm{VFM}(\rvx_0)$; intermediate feature $\rvf_t$ from $V_\theta(\rvz_t,t)$
  \State $\rvy_{\mathrm{latent}} = h_{\vomega_1}(\rvz_0); \rvy_{\mathrm{diff}} = h_{\vomega_2}(\rvf_t)$
  \State $ L_{\mathrm{REPA++}}=\mathcal L_{\mathrm{Latent-REPA}}(\rvy, \rvy_{\mathrm{latent}};\omega_1) + \mathcal L_{\mathrm{Diff-REPA}}(\rvy, \rvy_{\mathrm{diff}};\omega_2)$
  \State
  \Statex \emph{For }\textbf{{{\ours} (w/o REPA++)}}
  \State $L=L_{\mathrm{VAE}}+L_{\mathrm{Diff}}+L_{\mathrm{TPR}}$
  \Statex \emph{For }\textbf{{{\ours} (w/ REPA++)}}
  \State $L=L_{\mathrm{VAE}}+L_{\mathrm{Diff}}+L_{\mathrm{TPR}} + L_{\mathrm{REPA++}}$
  \State 
  \State Update $(\vphi,\vpsi,\vtheta)\leftarrow \mathrm{optimizer step on } \nabla L$
\EndWhile
\end{algorithmic}
\end{algorithm}

\subsubsection{Pipeline}\label{app:pipeline}
We summarize the training algorithm of {\ours} with pseudo-code in \cref{alg:main}. As visualized in \cref{fig:pipeline}, the algorithm consists of four loss branches: the VAE reconstruction loss (\cref{app:vae-loss}), the diffusion flow matching loss (\cref{app:diffusion-loss}), the Tweedie pixel reconstruction (TPR) loss (\cref{app:tpr-loss}), and the REPA++ alignment loss (\cref{app:repa++}).

\subsubsection{VAE Reconstruction}\label{app:vae-loss}

Given an image $\rvx_0$, the VAE encoder produces a Gaussian posterior
\begin{equation}
    p_{\vphi}(\rvz_0 \mid \rvx_0) = \mathcal N\big(\rvz_0; \vmu_{\vphi}(\rvx_0), \sigma_{\vphi}^2(\rvx_0)\mI\big),
\end{equation}
where $\vmu_{\vphi}(\rvx_0)$ and $\sigma_{\vphi}(\rvx_0)$ have the same dimensionality as the latent representation. The latent sample is obtained via the reparameterization trick:
\begin{equation}
    \rvz_0 = \vmu_{\vphi}(\rvx_0) + \sigma_{\vphi}(\rvx_0)\vepsilon,\quad \vepsilon \sim \mathcal N(\vzero, \mI).
\end{equation}
Although our VAE is not used solely for generative sampling as in classical VAEs, we find that retaining the variational formulation surprisingly benefits downstream diffusion training. This observation is further discussed in \cref{app:analysis-of-kl}. The reconstruction $\hat{\rvx}_{\text{vae}} = D_{\vpsi}(\rvz_0)$ is supervised by a combination of four losses: MSE, LPIPS, discriminator loss, and KL divergence.

\vspace{-10pt}
\paragraph{MSE Loss.}  
We compute the pixel-wise mean squared error between the clean and reconstructed images. Empirically, MSE and MAE yield similar visual quality, and we adopt MSE by default.

\vspace{-10pt}
\paragraph{LPIPS Loss.}  
We use the official LPIPS implementation~\cite{lpips} with a VGG backbone~\cite{vgg}. LPIPS is computed at the original image resolution without downsampling.

\vspace{-10pt}
\paragraph{Discriminator Loss.}
We incorporate a PatchGAN-style discriminator to enhance perceptual fidelity. To ensure stable adversarial training without relying on warm-up schedules or dynamically adjusted weights, we introduce two modifications:
\begin{itemize}
    \item Replace BatchNorm with SpectralNorm~\cite{miyato2018spectral} to improve Lipschitz continuity and reduce training instability.
    \item Apply LeCam regularization~\cite{lecam}, which tracks an exponential moving average of real and fake logits and penalizes large deviations, effectively smoothing discriminator updates.
\end{itemize}
These stabilizations allow us to adopt a fixed discriminator loss weight and enable adversarial supervision from the beginning of training without the need for special scheduling.

\vspace{-10pt}
\paragraph{Variational Objective (KL Loss).}  
The evidence lower bound (ELBO)~\cite{vae} has an additional KL divergence loss between the posterior $p_{\vphi}(\rvz_0|\rvx_0)$ and the standard Gaussian prior $\mathcal N(\vzero, \mI)$. Interestingly, the KL weight strongly affects the trade-off between reconstruction of VAE and generative of diffusion. As the KL weight increases, reconstruction performance decreases while generative quality initially improves and then degrades. We select the turning point that yields the best generation performance; details are provided in \cref{app:analysis-of-kl}.

\subsubsection{Diffusion Flow Matching}\label{app:diffusion-loss}

We adopt the linear flow-matching scheduler among several alternatives (variance preserving~\cite{ddpm}, variance exploding~\cite{sysde}, TrigFlow~\cite{scm}, etc.) due to its simplicity and strong empirical performance. The stochastic interpolant is defined as $\rvz_t = (1 - t)\rvz_0 + t\vepsilon$, where $\vepsilon \sim \mathcal N(\vzero, \mI)$.  
We use the velocity parameterization for diffusion model $V_\vtheta$, which predicts the vector field of the probability flow ODE. The flow-matching loss is
\begin{equation}
    \mathcal L_{\mathrm{Diff}}(\vphi, \vtheta)
= \mathbb E_{\rvz_0, \vepsilon, t}\!\left[
\|V_\vtheta(\rvz_t, t) - (\vepsilon - \rvz_0)\|_2^2
\right],
\end{equation}
where the time sampler for $t$ is specified in \cref{tab:detailed_config}. We observe that log-normal time sampling slightly accelerates Base model training, though the speedup becomes neglegible for Large and XL models.

\subsubsection{Tweedie Pixel Reconstruction (TPR) Loss}\label{app:tpr-loss}

For a given $\rvx_0$, sampled latent $\rvz_0$, and interpolant $\rvz_t$, we estimate the clean latent using Tweedie’s formula:
\begin{equation}
    \rvz_{\text{tweedie}} = \rvz_t - t\,V_\vtheta(\rvz_t, t).
\end{equation}
By definition, $\rvz_{\text{tweedie}} = \mathbb E[\rvz_0 \mid \rvz_t]$. Decoding this latent via the VAE decoder yields
\begin{equation}
    \hat{\rvx}_{\text{tpr}} = D_{\vpsi}(\rvz_{\text{tweedie}}).
\end{equation}
This provides an additional pixel-space anchor for the diffusion model. We apply a combination of MSE and LPIPS losses between $\hat{\rvx}_{\text{tpr}}$ and $\rvx_0$.

\subsubsection{REPA++}\label{app:repa++}

We adopt the REPA++ to align both the latent $\rvz_0$ and intermediate diffusion features $\rvf_t$ extracted from $\rvz_t$. The adapter is implemented as a two-layer MLP with SiLU activation.

\subsubsection{Loss Coefficients}

{\ours} consists of multiple loss components, yet we observe that the model is robust to a wide range of loss weights. Therefore, we adopt simple defaults: all losses use a weight of $1.0$ except the discriminator loss and KL loss, which are set to $0.1$ and $1\times10^{-5}$, respectively. The discriminator weight improves training stability, while the KL weight is selected based on the analysis in \cref{app:analysis-of-kl}.

\newpage
\subsection{Sampling Details}\label{app:sampling-detail}

In this section, we describe our guidance sampling strategy based on AutoGuidance~\cite{auto-guidance} and discuss how representation alignment affects Classifier-Free Guidance (CFG)~\cite{cfg} in {\ours}. In particular, we analyze why the gains from CFG tend to saturate when REPA++ is enabled.

\subsubsection{AutoGuidance}\label{app:guidance}

For all {\ours} variants, we train an additional \emph{bad} version of the diffusion model required by AutoGuidance. Concretely, this bad model is architecturally smaller: it uses half the number of transformer blocks and half the number of attention heads compared to the main model. To decouple it from end-to-end training, we pair this bad diffusion model with the frozen VAE obtained from the 200-epoch checkpoint and train the bad diffusion model from scratch for only 20 epochs.

At sampling time, we apply AutoGuidance using this bad model in conjunction with the main model. The guidance scale and guidance intervals are tuned per model size to achieve the best gFID score.

\subsubsection{Analysis of Unconditional Generation and CFG}\label{app:cfg-analysis}

We compare three settings for guidance: (1) without guidance, (2) CFG, and (3) AutoGuidance (AG). For {\ours} (w/ REPA++), we observe that the improvement from CFG over the unguided model is noticeably saturated, whereas AG continues to provide a substantial gain.

\begin{figure}[h!]
    \centering
    \includegraphics[width=\linewidth]{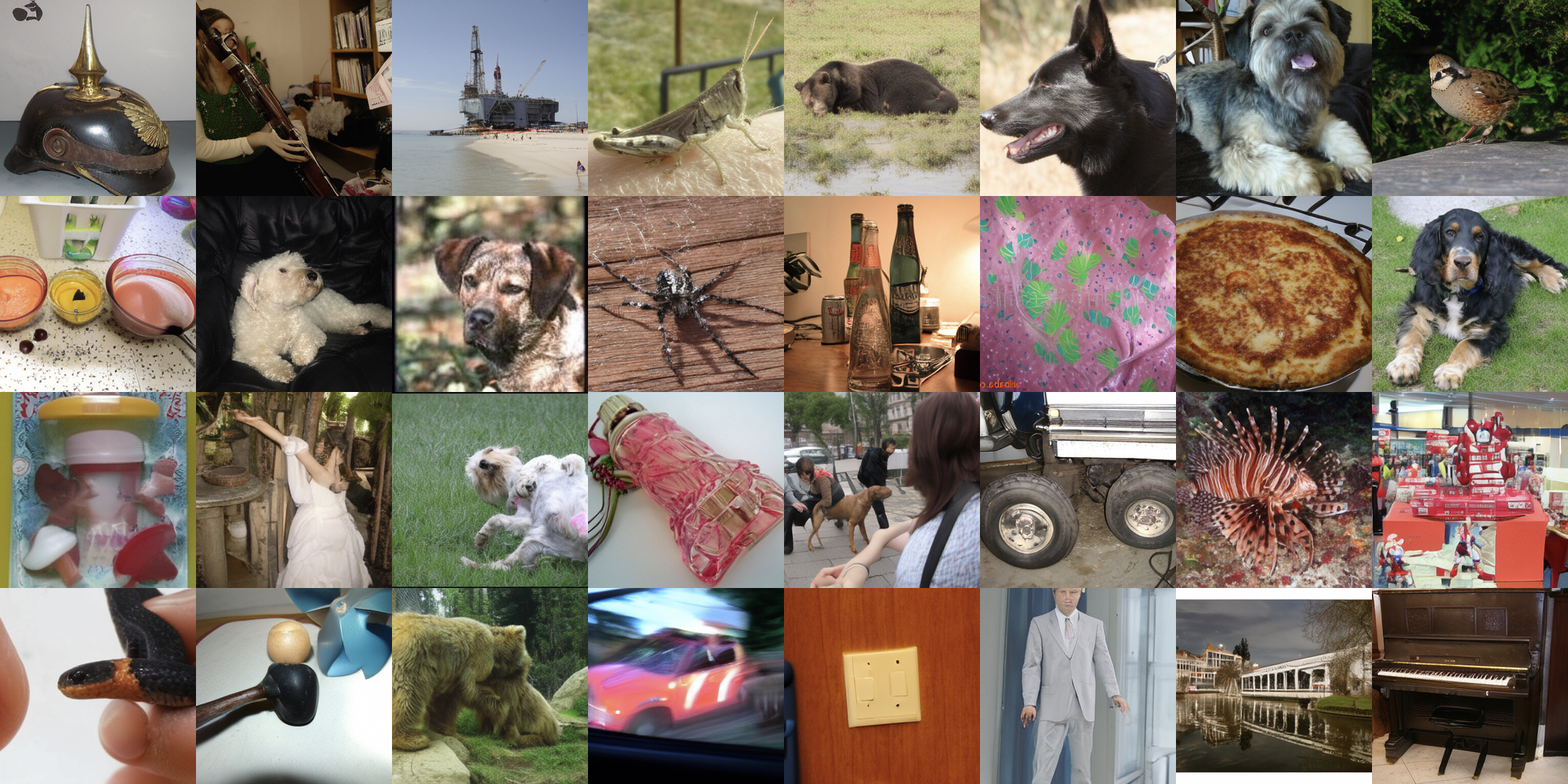}
    \caption{\textbf{Uncurated samples of unconditional generation.} On ImageNet 512$\times$512 with our {\ours}-XL (w/ REPA++) model (200 epochs) with the null-class. The unconditional generation is already surprisingly strong.}
    \label{fig:uncond-sample}
\end{figure}

After careful analysis, we identify a key contributing factor: the unconditional generation of {\ours} (with REPA++) is already surprisingly strong. In other words, the null-class (unconditional) predictions of the main model produce high-quality samples. We illustrate unconditional samples in \cref{fig:uncond-sample} and report their FID scores in \cref{tab:uncond-fid}. Due to the strong unconditional sampling quality, the negative direction in CFG provides only limited additional guidance. AutoGuidance avoids this issue by constructing the negative direction using the prediction of a deliberately degraded bad model under the conditional class. This alternative direction is substantially more informative, and as a result, AutoGuidance delivers a much larger improvement in FID compared to CFG.
\begin{table*}[h!]
\centering
\begin{minipage}{0.38\textwidth}
\centering
\caption{\textbf{Unconditional Generation.} gFID on ImageNet 256$\times$256 after 200 epochs.}

\label{tab:uncond-fid}
\renewcommand{\arraystretch}{1.12}
\setlength{\tabcolsep}{4pt}
\resizebox{\textwidth}{!}{%
\begin{tabular}{lc}
\toprule
\textbf{Method} & \textbf{gFID$\downarrow$ (unconditional)} \\
\midrule
{\ours}-XL (w/o REPA++) & 22.32 \\
{\ours}-XL (w/ REPA++)  & 4.03  \\
\bottomrule
\end{tabular}
}
\end{minipage}
\hfill
\begin{minipage}{0.6\textwidth}
\centering
\caption{\textbf{CFG vs.\ AG.} gFID on ImageNet 256$\times$256 after 200 epochs. CFG yields limited gains for {\ours} (w/ REPA++) due to its strong unconditional generation.}

\label{tab:guidance-fid}
\renewcommand{\arraystretch}{1.12}
\setlength{\tabcolsep}{4pt}
\resizebox{\textwidth}{!}{%
\begin{tabular}{lccc}
\toprule
\textbf{Method} & \textbf{gFID$\downarrow$ (w/o guidance)} & \textbf{gFID$\downarrow$ (w/ CFG)} & \textbf{gFID$\downarrow$ (w/ AG)}\\
\midrule
{\ours}-XL (w/o REPA++) & 2.42 & 1.52 & 1.46 \\
{\ours}-XL (w/ REPA++)  & 1.50 & 1.45 & 1.21 \\
\bottomrule
\end{tabular}
}
\end{minipage}
\end{table*}

\newpage
\definecolor{rowhl}{HTML}{E8F3FF}
\begin{table*}[t]
\centering
\renewcommand{\arraystretch}{1.12}
\setlength{\tabcolsep}{4pt}
\caption{\textbf{ImageNet 256$\times$256 generation benchmark (continue).} 
We report additional results for using different VAE and diffusion architectures to demonstrate the superior performance of refined architecture for both non–representation alignment and representation alignment methods ($\downarrow$ lower is better; $\uparrow$ higher is better).}
\label{tab:archi}
\resizebox{\textwidth}{!}{%
\begin{tabular}{l!{\vline}l!{\vline}l!{\vline}l!{\vline}l!{\vline}c!{\vline}*{5}{c}!{\vline}*{5}{c}c}
\toprule
\multirow{2}{*}{\textbf{Method}} &
\multicolumn{2}{c!{\vline}}{\textbf{VAE}} &
\multicolumn{2}{c!{\vline}}{\textbf{Diffusion}} &
\multirow{2}{*}{\textbf{Epochs}} &
\multicolumn{5}{c!{\vline}}{\textbf{Generation w/o guidance}} &
\multicolumn{5}{c}{\textbf{Generation w/ guidance}} &
\multirow{2}{*}{\textbf{rFID$\downarrow$}} \\
\cmidrule(lr){2-3}\cmidrule(lr){4-5}\cmidrule(lr){7-11}\cmidrule(lr){12-16}
& \textbf{Model} & \textbf{Params (M)} & \textbf{Model} & \textbf{Params (M)} & &
\textbf{gFID$\downarrow$} & \textbf{sFID$\downarrow$} & \textbf{IS$\uparrow$} & \textbf{Prec.$\uparrow$} & \textbf{Rec.$\uparrow$} &
\textbf{gFID$\downarrow$} & \textbf{sFID$\downarrow$} & \textbf{IS$\uparrow$} & \textbf{Prec.$\uparrow$} & \textbf{Rec.$\uparrow$} & \\
\midrule
\midrule
\rowcolor{rowhl}
\multicolumn{17}{c}{\textbf{Non-representation Alignment Methods}} \\
\midrule
\midrule
\multirow{7}{*}{{\ours}} & CNN-VAE & 70 & DiT-XL/1 & 675 & 80  & 5.89  & 5.49 & 145.58  & 0.75  & 0.61 & 2.25 & 5.35 & 254.22 & 0.81 & 0.57 & 0.73 \\
\cmidrule(lr){2-17}
\multirow{7}{*}{\small{(w/o REPA++)}}& \multirow{3}{*}{CNN-VAE} & \multirow{3}{*}{70} & {{\ourdit}-B} & 91 & 80 & 9.24   & 5.53   & 111.26  & 0.72  & 0.59 & 3.50 & 5.23 & 200.70 & 0.80 & 0.54 & 0.69\\
&  &  & {{\ourdit}-L} & 319 & 80 & 4.62   & 4.53   & 150.40  & 0.77  & 0.59 & 2.17 & 4.75 & 251.28 & 0.82 & 0.56 & 0.67\\
&  &  & {{\ourdit}-XL} & 672 & 80 & 3.71   & 4.53   & 166.52  & 0.78  & 0.61 & 2.12 & 4.69 & 252.60 & 0.83 & 0.56 & 0.61\\
\cmidrule(lr){2-17}

& {\ourvae}-B & 58 & {\ourdit}-B  & 91 & 80 & 10.65   & 5.67   & 105.59  & 0.71  & 0.59 & 3.86 & 5.13 & 198.07 & 0.81 & 0.54 & 0.74 \\
& {\ourvae}-L & 207 & {\ourdit}-L  & 319 & 80 & 5.31   & 4.78   & 144.62  & 0.76  & 0.61 & 2.15 & 4.65 & 245.47 & 0.82 & 0.58 & 0.58 \\
 & {\ourvae}-XL & 437 & {\ourdit}-XL  & 672 & 80 & 3.66  & 4.42 & 172.03    & 0.78  & 0.62 & 1.80 & 4.40 & 267.10 & 0.81 & 0.59 & 0.54 \\
 
\midrule
\midrule
\rowcolor{rowhl}
\multicolumn{17}{c}{\textbf{Representation Alignment Methods}} \\
\midrule
\midrule
\multirow{7}{*}{{\ours}} & CNN-VAE & 70 & DiT-XL/1 & 675 & 80  & 2.83   & 4.31  & 185.36  & 0.78  & 0.62 & 1.80 & 4.37 & 273.49 & 0.79 & 0.63 & 0.62 \\
\cmidrule(lr){2-17}
\multirow{7}{*}{(w/ REPA++)} & \multirow{3}{*}{CNN-VAE} & \multirow{3}{*}{70} & {{\ourdit}-B} & 91 & 80 & 5.82   & 5.08   & 142.07  & 0.75  & 0.60 & 2.78 & 5.28 & 216.66 & 0.78 & 0.58 & 0.62 \\
& & & {{\ourdit}-L} & 319 & 80 & 2.65   & 4.33   & 183.23  & 0.79  & 0.61 & 1.75 & 4.66 & 271.58 & 0.80 & 0.61 & 0.60 \\
&  &  & {{\ourdit}-XL} & 672 & 80 & 2.15 & 4.21 & 206.49  & 0.80 & 0.62  & 1.74 & 4.55 & 279.13 & 0.81 & 0.61 & 0.64 \\
\cmidrule(lr){2-17}

& {\ourvae}-B & 58 & {\ourdit}-B  & 91 & 80 & 6.20 & 5.12 & 138.71  & 0.74  & 0.60 & 2.29 & 4.83 & 230.33 & 0.79 & 0.57 & 0.84\\
& {\ourvae}-L & 207 & {\ourdit}-L  & 319 & 80 & 2.15   & 4.21   & 207.87  & 0.79  & 0.61 & 1.41 & 4.15 & 292.90 & 0.80 & 0.61 & 0.62 \\
 & {\ourvae}-XL & 437 & {\ourdit}-XL  & 672 & 80 & 1.69   & 4.14   & 234.56  & 0.80  & 0.63 & 1.29 & 4.08 & 309.42 & 0.80 & 0.63 & 0.59 \\

\bottomrule
\end{tabular}
}%
\end{table*}

\section{Additional Results and Ablations}\label{app:add-results}
\subsection{Joint Scaling of VAE and Diffusion}\label{app:joint-scaling}

In this section, we examine how jointly scaling the VAE and the diffusion model affects performance. Unlike prior latent-diffusion pipelines that pair a fixed-size VAE with diffusion models of various capacities, {\ours} benefits substantially from joint scaling both components. A larger VAE not only produces a more expressive latent space but also provides stronger feedback to the diffusion model during end-to-end training.

\begin{figure*}[h!]
    \centering
    \includegraphics[width=\linewidth]{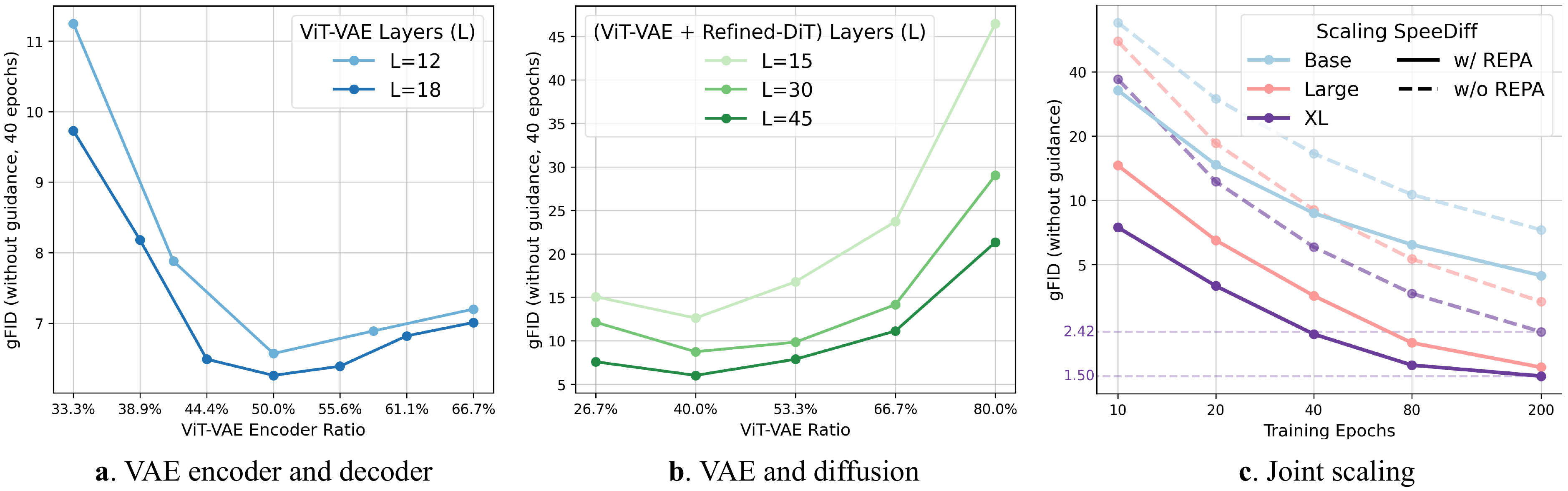}
    \caption{\textbf{Joint scaling of VAE and diffusion.} On ImageNet 256$\times$256 generation, we examine how to allocate transformer layers between encoder, decoder, and diffusion. (a) Encoder and decoder perform best with a balanced 1:1 ratio. (b) Allocating about 40\% of total layers to the VAE gives a strong trade-off. (c) A 1:1:3 allocation (encoder:decoder:diffusion) consistently works well with and without REPA++.}
    \label{fig:scaling}
\end{figure*}

To understand how model capacity should be allocated, we systematically vary the number of transformer layers assigned to the encoder, decoder, and diffusion model. As shown in \cref{fig:scaling}, several trends emerge. First, the encoder and decoder perform best when their depths are approximately balanced (around a 1:1 ratio). Second, dedicating roughly 40\% of total layers to the VAE yields a strong trade-off between reconstruction quality and downstream generative performance. Finally, we observe a consistent and stable scaling trend across model sizes, both with and without REPA++, when following a 1:1:3 allocation rule for encoder, decoder, and diffusion layers.

\subsection{{\ours} with Conventional Architecture}

We further evaluate the impact of our refined architectures by comparing {\ourvae} and {\ourdit} with conventional CNN-VAE and DiT variants. As shown in \cref{tab:archi}, both refinements consistently improve performance across model scales. 

Across both non-aligned and REPA++ settings, our refined architectures provide consistent improvements over conventional CNN-VAE and DiT designs. Without alignment, replacing DiT with {\ourdit} already yields substantial gains (e.g., gFID improves from 5.89 to 3.71 with CNN-VAE), and replacing CNN-VAE with {\ourvae} further enhances performance, reaching 3.66 at the XL scale on ImageNet 256$\times$256 generation after 80 epochs. With REPA++, the benefits become even more pronounced: {\ourdit}-XL lowers gFID from 2.83 to 2.15, and jointly scaling the VAE achieves the best overall result of 1.69 on ImageNet 256$\times$256 generation after 80 epochs. Together, these findings confirm the effectiveness of our refined architectures.

\subsection{Analysis the Effect of KL Loss}\label{app:analysis-of-kl}

The variational objective (KL loss) in the VAE loss plays a crucial role in balancing reconstruction of VAE and generative of diffusion. Recall that the KL term measures the divergence between the encoder posterior $p_{\vphi}(\rvz_0 \mid \rvx_0)$ and the standard Gaussian prior $\mathcal N(\vzero, \mI)$:
\begin{equation}
D_{\mathrm{KL}}\!\left(
\mathcal N\big(\vmu_{\vphi}, \sigma_{\vphi}^2 \mI\big)
\;\big\|\;
\mathcal N(\vzero, \mI)
\right)
= 
\frac{1}{2}
\Bigg[
\underbrace{\sum_{i=1}^d \vmu_{\vphi,i}^2}_{\text{mean penalty}}
\;+\;
\underbrace{\sum_{i=1}^d \left(\sigma_{\vphi,i}^2 - \log \sigma_{\vphi,i}^2 - 1 \right)}_{\text{variance penalty}}
\Bigg].
\end{equation}
Through extensive experiments, we find that the weight assigned to this KL term has a surprisingly strong influence on downstream diffusion performance. Setting the KL weight to zero effectively turns the model into a standard autoencoder (AE). Although this preserves reconstruction quality, it consistently harms generative performance, suggesting that a weakly regularized latent space is less suitable for diffusion training. Conversely, assigning too large a KL weight forces the latent distribution toward an overly restrictive prior, degrading both reconstruction and generation. Empirically, we observe that the average predicted standard deviation $\sigma_{\vphi,i}$ increases monotonically as the KL weight is raised from zero to sufficiently large values. This trend is summarized in \cref{tab:kl-loss}, based on which we select the turning point of the KL weight that provides the best generative performance.

\begin{table}[h!]
\centering
\caption{\textbf{Effect of KL Loss.} We study how the KL loss shapes the latent space by measuring the average latent log-variance $\log\sigma^2_{\vphi}$ across spatial locations and samples. We additionally report both generation and reconstruction metrics on ImageNet 256$\times$256 after 80 epochs with {\ours}-XL (w/ REPA++).}

\label{tab:kl-loss}
\renewcommand{\arraystretch}{1.1}
\setlength{\tabcolsep}{6pt}
\resizebox{0.6\textwidth}{!}{%
\begin{tabular}{c!{\vline}cccccc}
\toprule
\textbf{KL weight} & 0 (AE) & 1e-6 & 3e-6 & 1e-5 (\emph{best}) & 3e-5 & 1e-4 \\   %
\midrule
\midrule
\textbf{Average log-variance} & -123.31  & -13.26 &  -9.25 & -6.87 & -4.31 & -2.13\\
\midrule
\textbf{gFID$\downarrow$} & 3.91 & 2.48 & 1.94 & \hldigit{1.69} & 1.86 & 2.76 \\
\textbf{rFID$\downarrow$} & \hldigit{0.34} & 0.39 & 0.44 & 0.59 & 0.87 & 1.46\\ 
\textbf{PSNR$\uparrow$} & \hldigit{27.59} & 27.33 & 26.98 & 26.05 & 24.83 & 22.08\\
\bottomrule
\end{tabular}
}
\end{table}

\subsection{Sensitivity to TPR Loss Weight}

We ablate the TPR loss weight to examine the sensitivity of {\ours} to this hyperparameter. As shown in \cref{tab:tpr_weight_ablation_transposed}, performance remains stable across a wide range of weights (0.5--4.0), with all settings yielding comparable gFID, rFID, and PSNR. Notably, even a $4\times$ increase from the default weight of 1.0 causes negligible degradation in generation quality (gFID 6.33 vs.\ 6.20), while slightly improving reconstruction metrics. This robustness indicates that {\ours} does not require careful tuning of the TPR loss weight in practice.

\begin{table}[htbp]
\caption{Ablation on TPR loss weight with {\ours}-B (w/ REPA++) on ImageNet $256\times256$ after 80 epochs.}
\label{tab:tpr_weight_ablation_transposed}
\centering
\resizebox{0.5\linewidth}{!}{%
\begin{tabular}{lccccc}
\toprule
\textbf{TPR weight} & \textbf{0.0} & \textbf{0.5} & \textbf{1.0 (default)} & \textbf{2.0} & \textbf{4.0} \\
\midrule
gFID$\downarrow$ & 12.69 & 6.91 & 6.20 & \hldigit{6.08} & 6.33 \\
rFID$\downarrow$ &   0.90  &  0.85    & 0.84 &   0.80   &  \hldigit{0.74}    \\
PSNR$\uparrow$   &  25.40     &  25.56   &  25.61    &  25.65    &  \hldigit{25.87}    \\
\bottomrule
\end{tabular}
}
\end{table}

\subsection{Analysis of Practical Cost}
\label{subsec:practical_cost}

We analyze the practical cost of {\ours} by measuring the time-level speedup attributable to end-to-end training. Specifically, we compare {\ours} (w/o REPA++) against the two-stage baseline, which assumes a pretrained VAE with preprocessed latents and thus requires only \emph{a single diffusion forward pass per training step}. In contrast, our end-to-end approach additionally involves \emph{one VAE encoder call and two decoder calls per step}, resulting in an approximately 2$\times$ slower per-step time (see \cref{tab:xl_comparison} for detailed per-step timing and \cref{tab:model_config} for model FLOPs).

Despite this per-step overhead, {\ours} converges dramatically faster in terms of training steps. Taking the baseline's 7.93 FID without guidance at 4M steps as the reference point, {\ours} reaches the same FID in only 0.17M steps. This translates to a 23.52$\times$ step-level speedup and, after accounting for the per-step cost difference, a 12.14$\times$ time-level speedup. Notably, these numbers are conservative: the two-stage pipeline additionally requires VAE pretraining or finetuning as a prerequisite, whereas {\ours} trains the VAE jointly from scratch, further widening the practical efficiency gap.

\begin{table}[htbp]
\caption{\textbf{Practical cost and speedup}. We evaluate on ImageNet 256$\times$256 with 16 H100-SXM-80GB GPUs at batch size 256.}
\label{tab:xl_comparison}
\centering
\resizebox{0.7\linewidth}{!}{
\begin{tabular}{lcccc}
\toprule
 & Step time (s) & Required steps (M) & Step-level speedup & Time-level speedup \\
\midrule
Two-stage-XL & 0.16 & 4 & -- & -- \\
{\ours}-XL    & 0.31 & 0.17 & $23.52\times$ & $12.14\times$ \\
\bottomrule
\end{tabular}}
\end{table}

Regarding memory consumption, {\ours}-XL requires 57.34 GB peak memory compared to 26.97 GB for the two-stage baseline under the same training configuration. While the higher memory footprint is a limitation of end-to-end training, we consider it a reasonable trade-off given the substantial convergence speedup achieved.

\section{Additional Qualitative Samples}\label{app:sample}
\begin{figure}[htbp]
    \centering
    \includegraphics[width=0.929\linewidth]{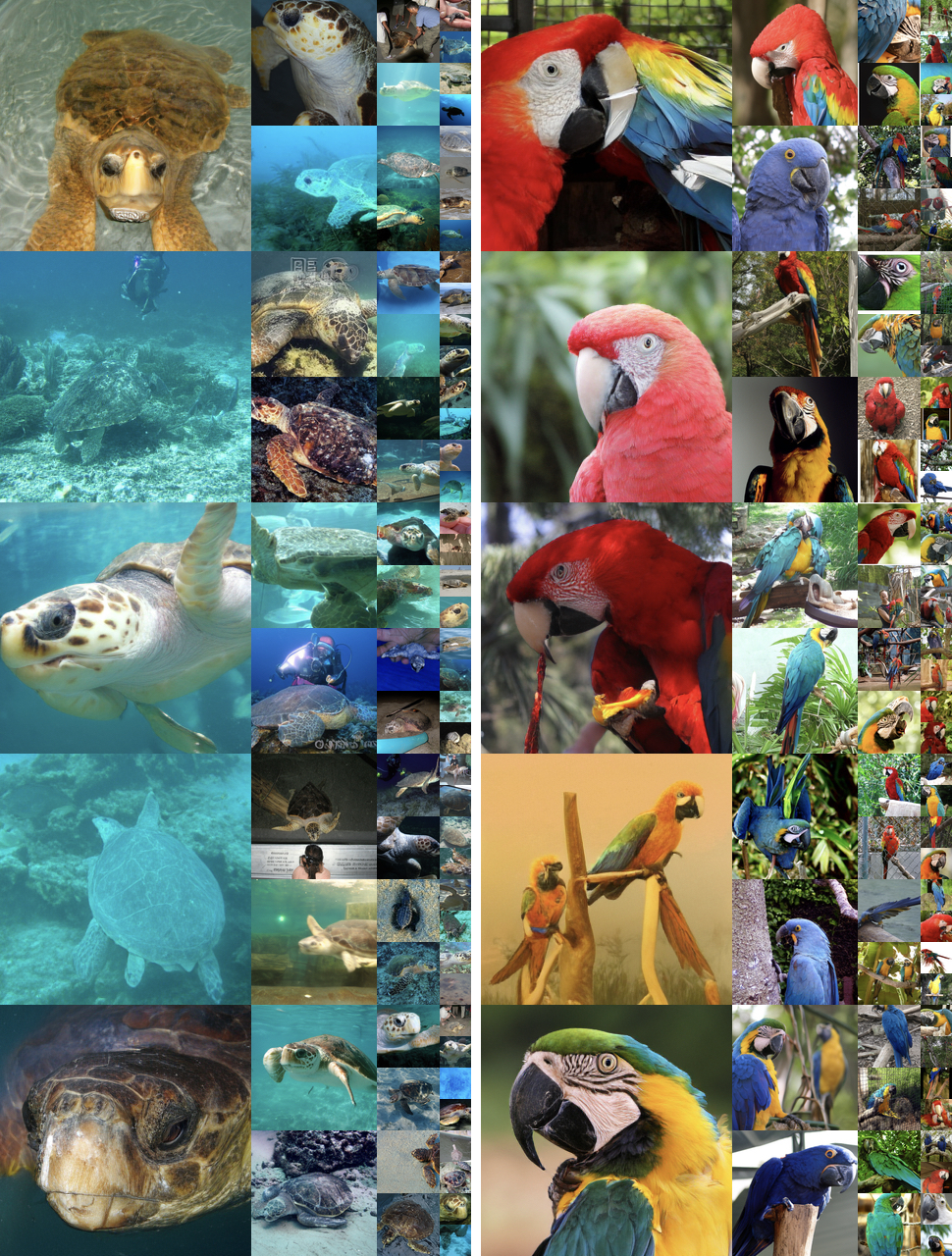}
    \caption{\textbf{Uncurated samples without guidance.} On ImageNet 256$\times$256  with our {\ours}-XL (w/ REPA++) model (200 epochs). The  class labels are \emph{loggerhead} (33, left) and \emph{macaw} (88, right).}
    \label{fig:app-v1}
\end{figure}

\begin{figure}[t]
    \centering
    \includegraphics[width=0.929\linewidth]{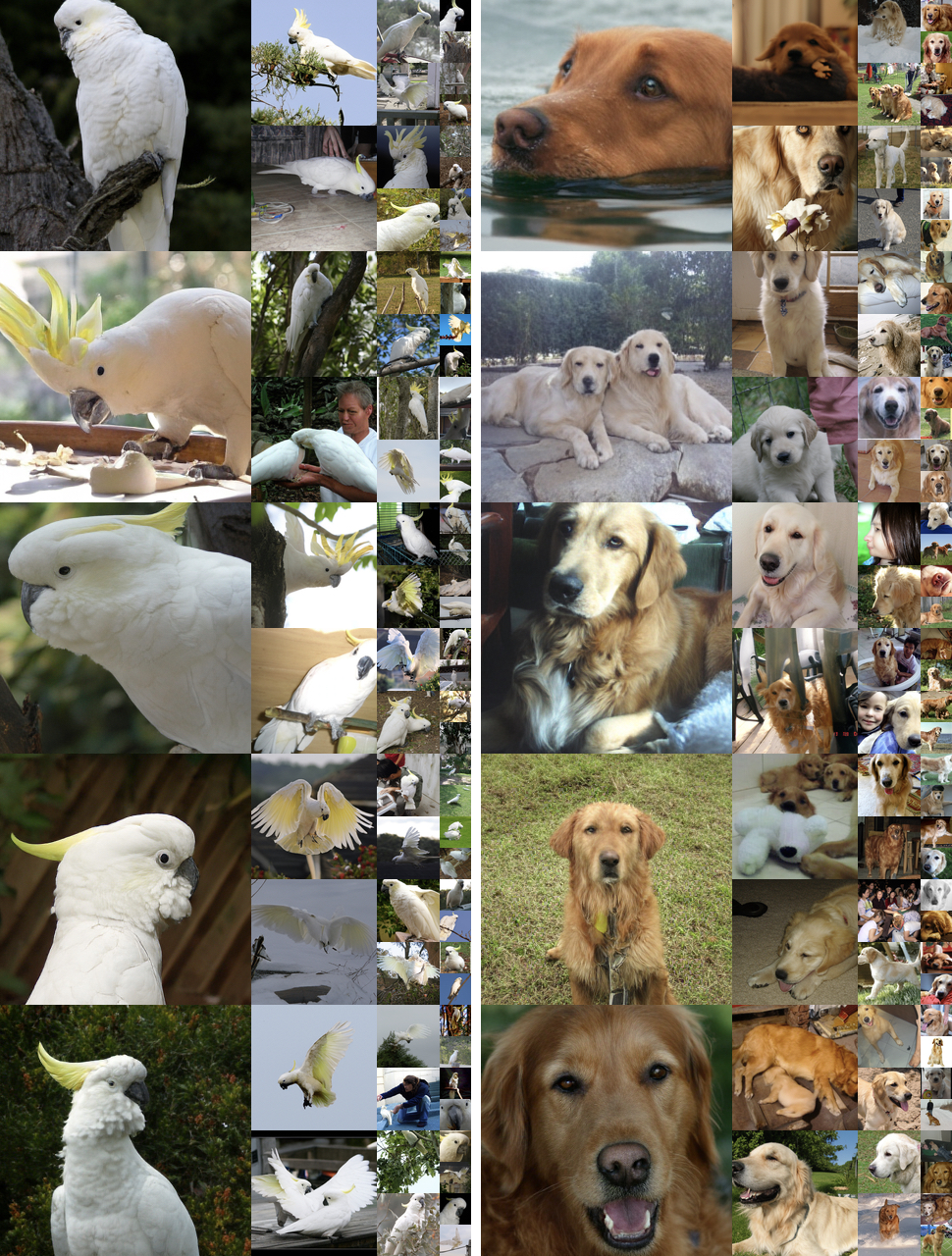}
    \caption{\textbf{Uncurated samples without guidance.} On ImageNet 256$\times$256  with our {\ours}-XL (w/ REPA++) model (200 epochs). The  class labels are \emph{sulphur-crested cockatoo} (89, left) and \emph{golden retriever} (207, right).}
    \label{fig:app-v2}
\end{figure}

\begin{figure}[htbp]
    \centering
    \includegraphics[width=0.929\linewidth]{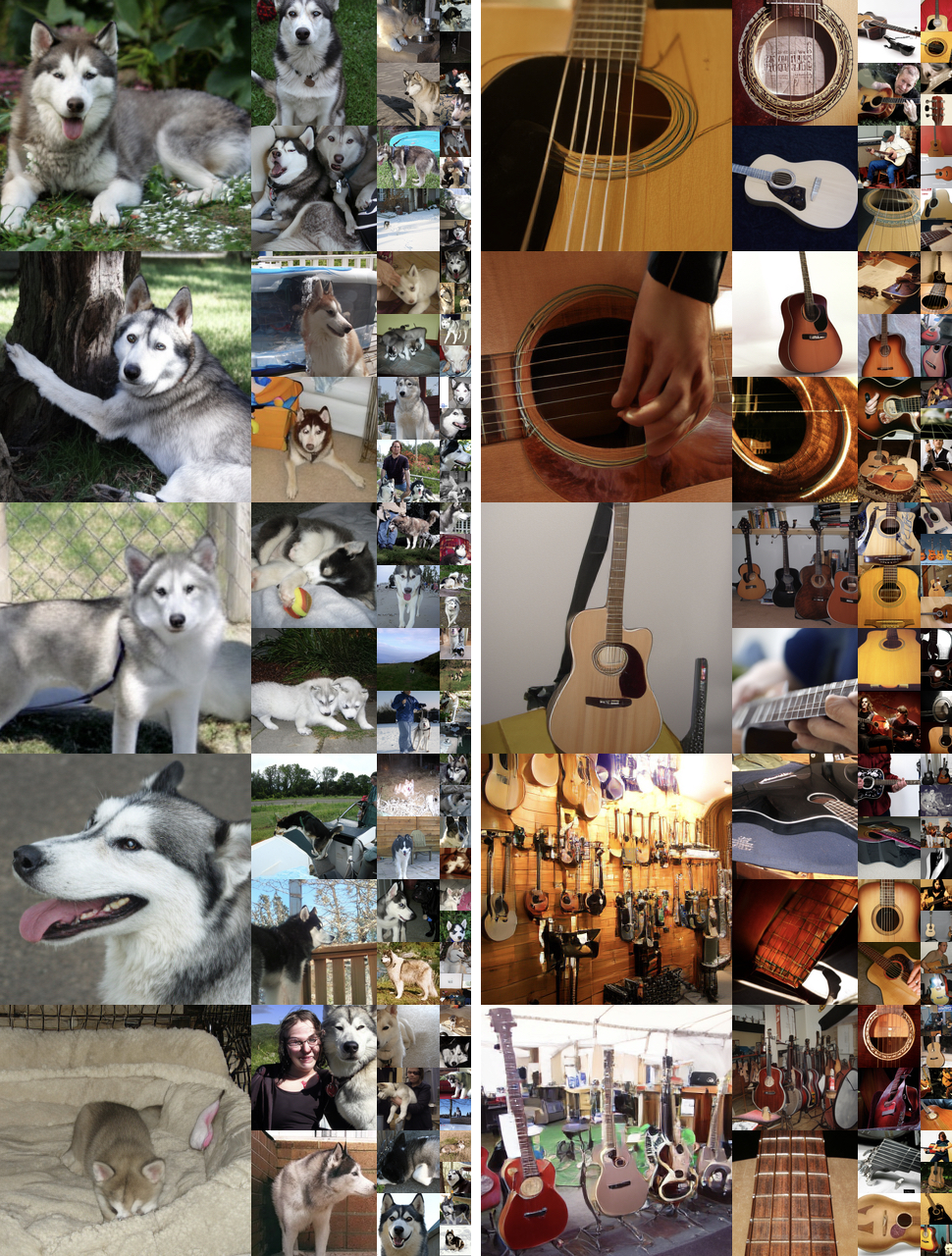}
    \caption{\textbf{Uncurated samples without guidance.} On ImageNet 256$\times$256  with our {\ours}-XL (w/ REPA++) model (200 epochs). The  class labels are \emph{Siberian husky} (250, left) and \emph{acoustic guitar} (402, right).}
    \label{fig:app-v3}
\end{figure}

\begin{figure}[htbp]
    \centering
    \includegraphics[width=0.929\linewidth]{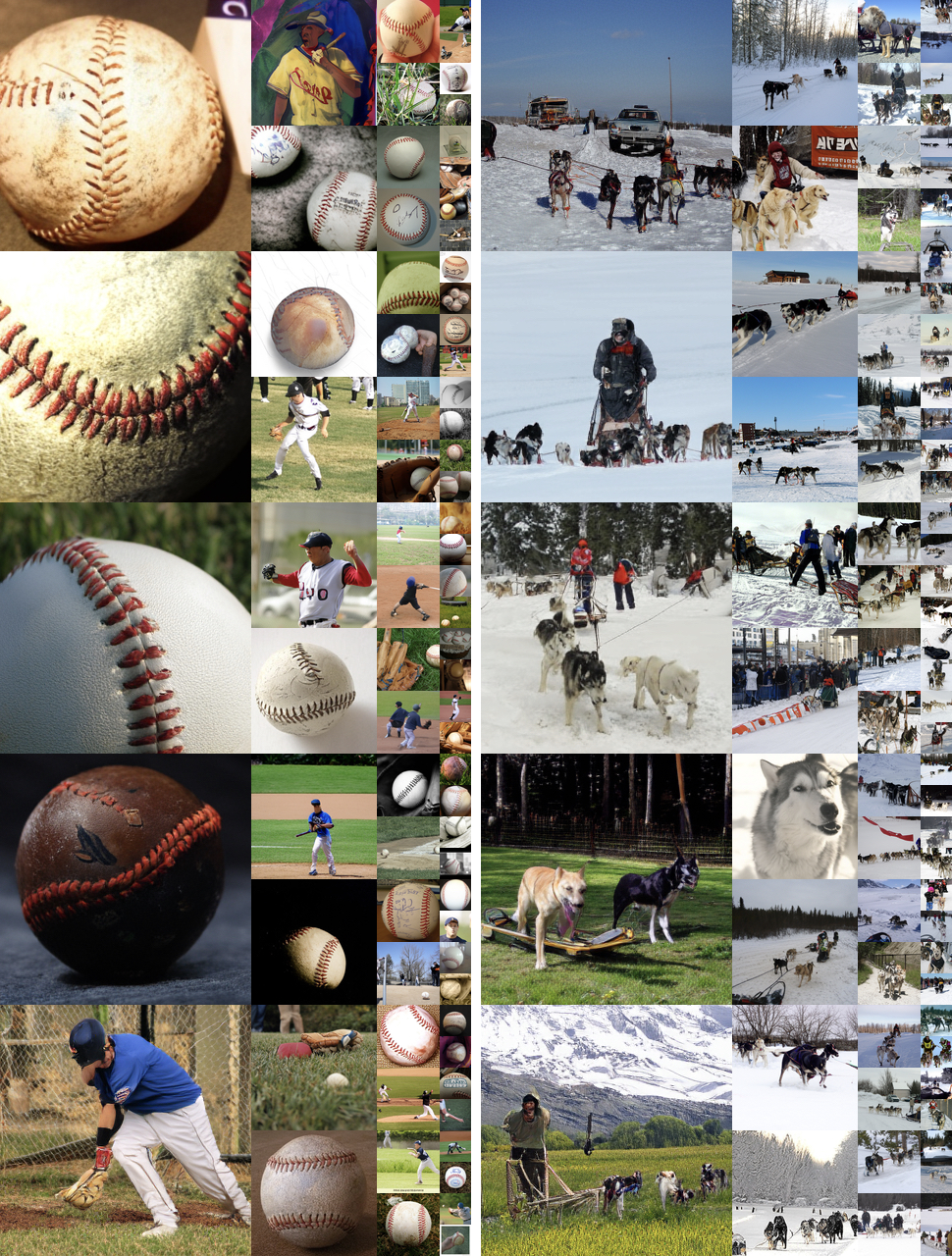}
    \caption{\textbf{Uncurated samples without guidance.} On ImageNet 256$\times$256  with our {\ours}-XL (w/ REPA++) model (200 epochs). The  class labels are \emph{baseball} (429, left) and \emph{dogsled} (537, right).}
    \label{fig:app-v4}
\end{figure}

\begin{figure}[htbp]
    \centering
    \includegraphics[width=0.929\linewidth]{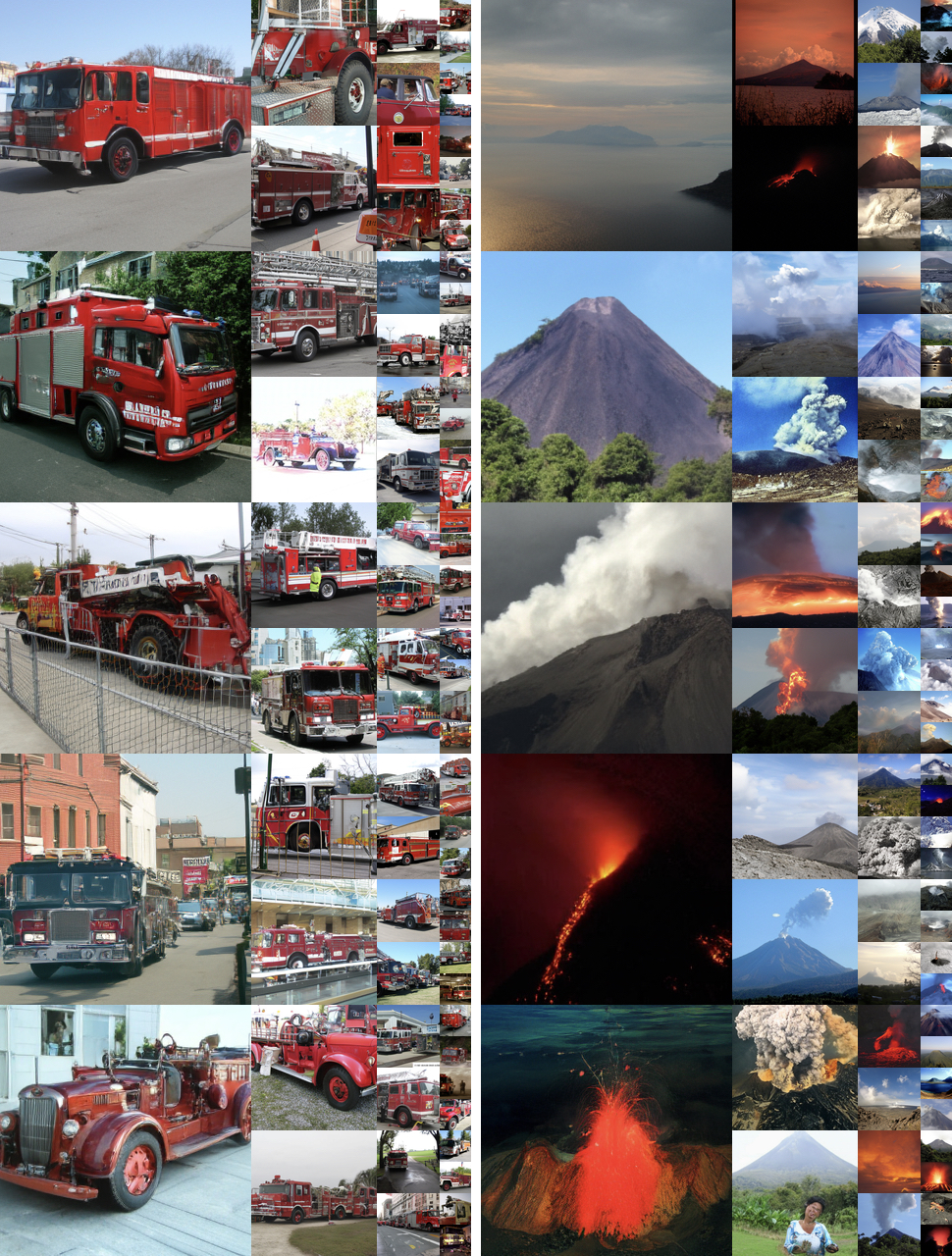}
    \caption{\textbf{Uncurated samples without guidance.} On ImageNet 256$\times$256  with our {\ours}-XL (w/ REPA++) model (200 epochs). The  class labels are \emph{fire truck} (555, left) and \emph{volcano} (980, right).}
    \label{fig:app-v5}
\end{figure}

\begin{figure}[htbp]
    \centering
    \includegraphics[width=0.929\linewidth]{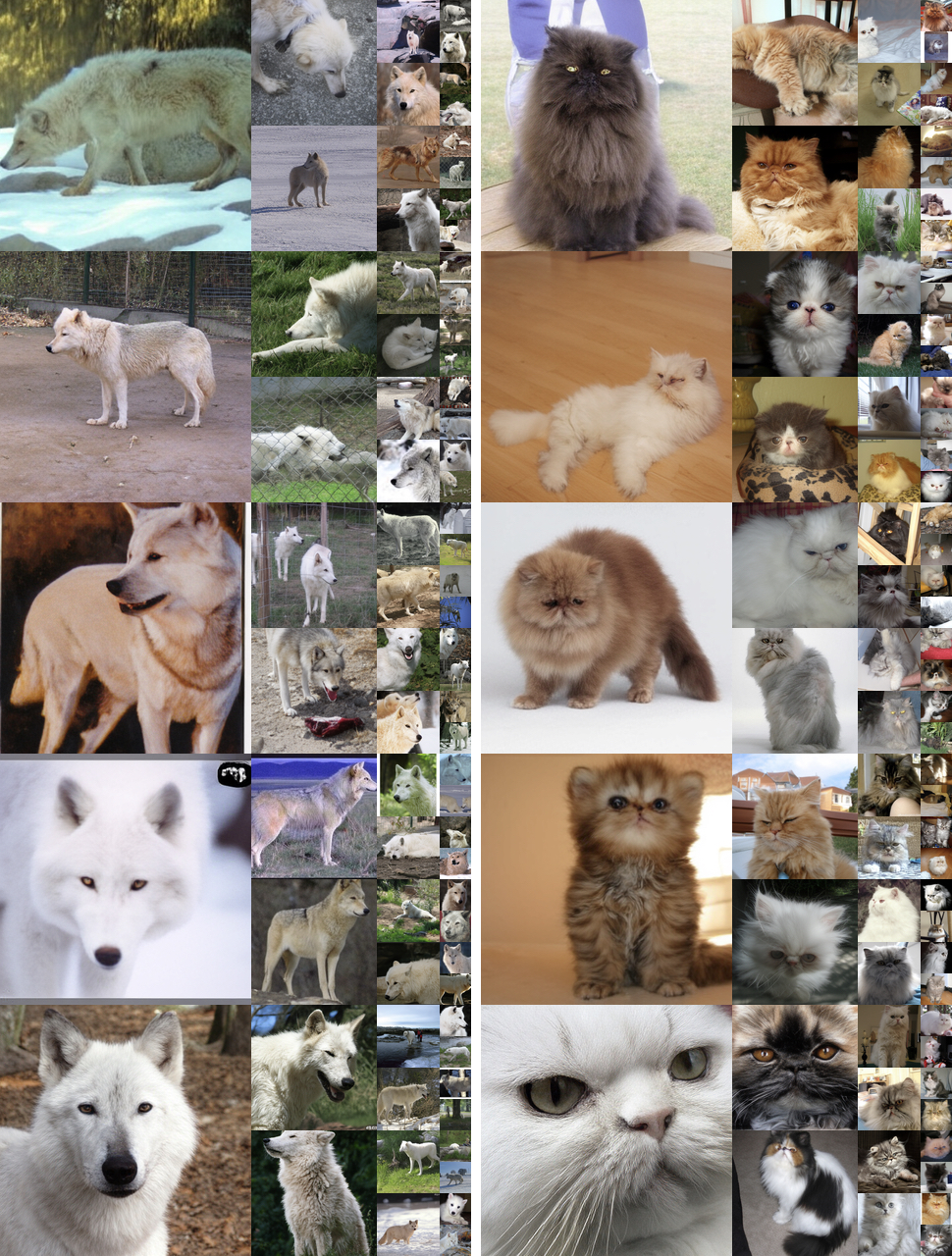}
    \caption{\textbf{Uncurated samples without guidance.} On ImageNet 512$\times$512  with our {\ours}-XL (w/ REPA++) model (200 epochs). The  class labels are \emph{white wolf} (270, left) and \emph{Persian cat} (283, right).}
    \label{fig:app-v1-512}
\end{figure}

\begin{figure}[htbp]
    \centering
    \includegraphics[width=0.929\linewidth]{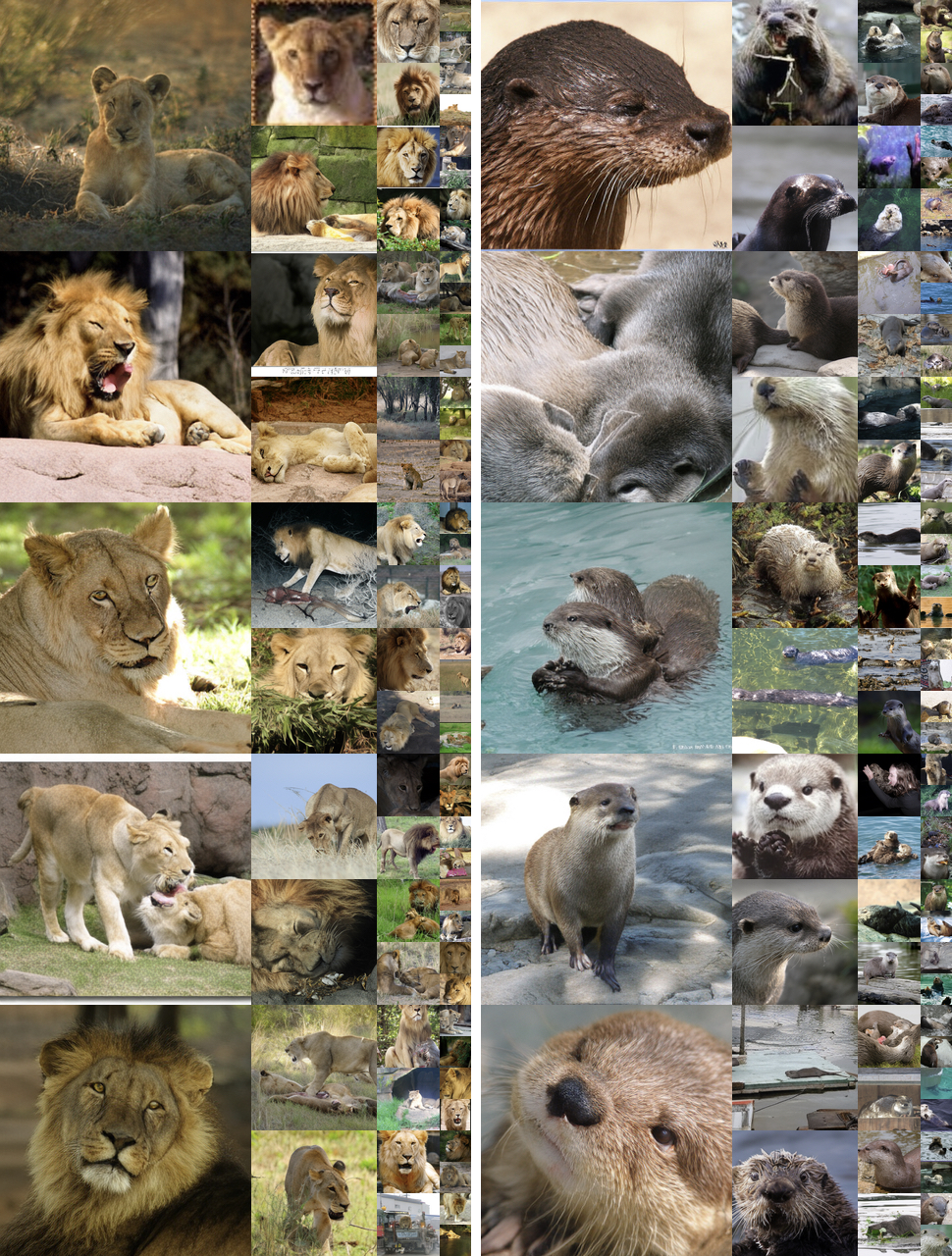}
    \caption{\textbf{Uncurated samples without guidance.} On ImageNet 512$\times$512  with our {\ours}-XL (w/ REPA++) model (200 epochs). The  class labels are \emph{lion} (291, left) and \emph{otter} (360, right).}
    \label{fig:app-v2-512}
\end{figure}

\begin{figure}[htbp]
    \centering
    \includegraphics[width=0.929\linewidth]{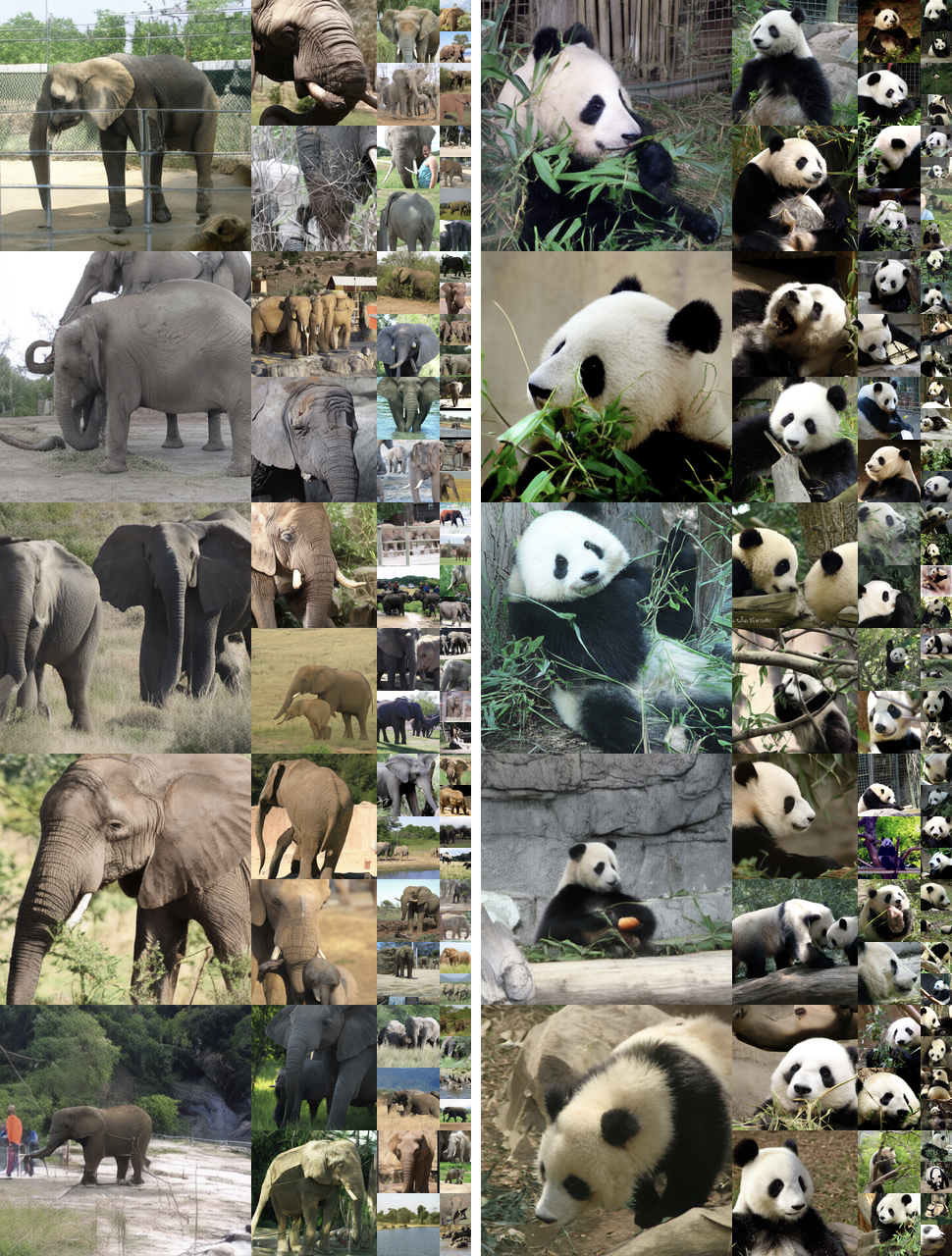}
    \caption{\textbf{Uncurated samples without guidance.} On ImageNet 512$\times$512  with our {\ours}-XL (w/ REPA++) model (200 epochs). The  class labels are \emph{African elephant} (386, left) and \emph{panda} (388, right).}
    \label{fig:app-v3-512}
\end{figure}

\begin{figure}[htbp]
    \centering
    \includegraphics[width=0.929\linewidth]{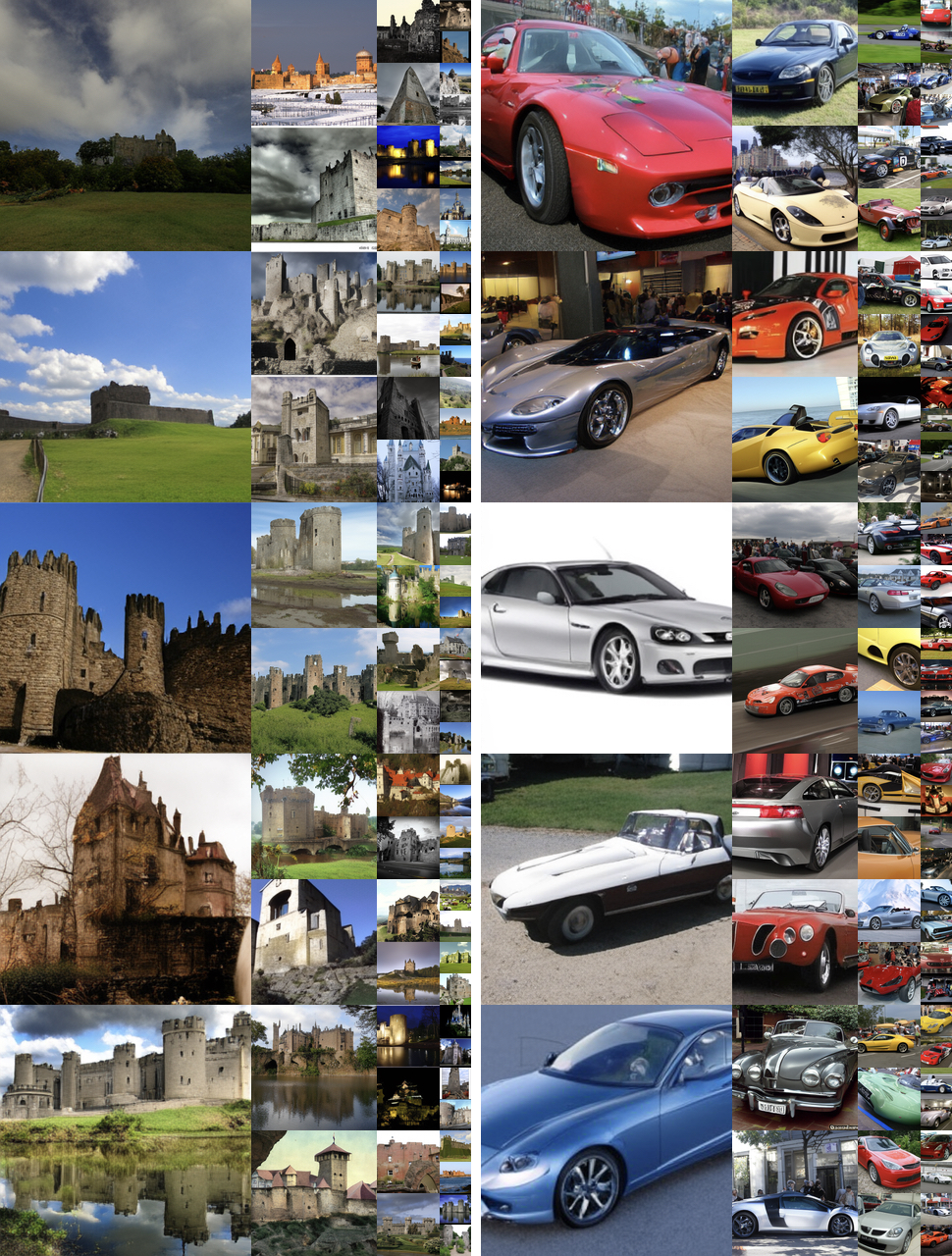}
    \caption{\textbf{Uncurated samples without guidance.} On ImageNet 512$\times$512  with our {\ours}-XL (w/ REPA++) model (200 epochs). The  class labels are \emph{castle} (483, left) and \emph{sports car} (817).}
    \label{fig:app-v4-512}
\end{figure}

\begin{figure}[htbp]
    \centering
    \includegraphics[width=0.929\linewidth]{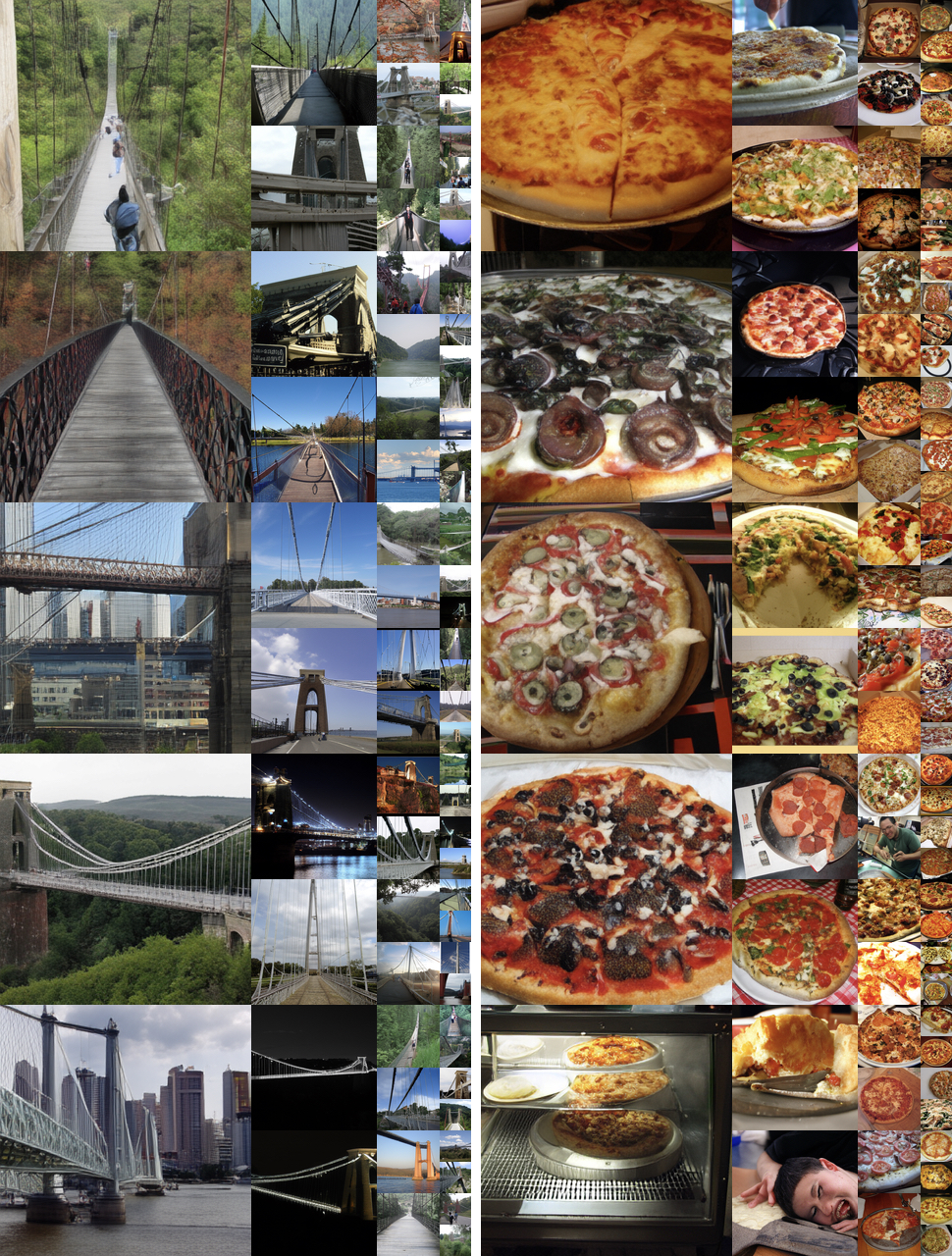}
    \caption{\textbf{Uncurated samples without guidance.} On ImageNet 512$\times$512  with our {\ours}-XL (w/ REPA++) model (200 epochs). The  class labels are \emph{suspension bridge} (839, left) and \emph{pizza} (963).}
    \label{fig:app-v5-512}
\end{figure}

We show additional samples of {\ours}-XL on both ImageNet 256$\times$256 and 512$\times$512 generation in \cref{fig:app-v1,fig:app-v2,fig:app-v3,fig:app-v4,fig:app-v5} and \cref{fig:app-v1-512,fig:app-v2-512,fig:app-v3-512,fig:app-v4-512,fig:app-v5-512}, respectively.
\begin{table*}[htbp]
\centering
\renewcommand{\arraystretch}{1.12}
\setlength{\tabcolsep}{4pt}
\caption{\textbf{Detailed implementation of {\ours} on ImageNet at 256$\times$256 and 512$\times$512 resolutions.} We report architecture, loss functions, training, and sampling hyperparameters. Rows highlighted in \colorbox{diffhl}{gray} indicate values that differ across models.}
\label{tab:detailed_config}
\resizebox{\textwidth}{!}{%
\begin{tabular}{lcccccc}
\toprule
 & \textbf{{\ours}-B} (256) & \textbf{{\ours}-L} (256) & \textbf{{\ours}-XL} (256) & \textbf{{\ours}-B} (512) & \textbf{{\ours}-L} (512) & \textbf{{\ours}-XL} (512)\\
\midrule\midrule
\multicolumn{4}{l}{\textbf{{\ourvae} details}} \\
\rowcolor{diffhl}
Size & Base & Large & XL & Base & Large & XL \\
\rowcolor{diffhl}
Patch size & 16 & 16 & 16 & 32 & 32 & 32 \\
\rowcolor{diffhl}
Layers & 4 & 8 & 9 & 4 & 8 & 9 \\
\rowcolor{diffhl}
Hidden dims & 768 & 1024 & 1408 & 768 & 1024 & 1408 \\
\rowcolor{diffhl}
Numbers of Heads & 12 & 16 & 22 & 12 & 16 & 22 \\
Dims per heads & 64 & 64 & 64 & 64 & 64 & 64 \\
Latent channels & 32 & 32 & 32 & 32 & 32 & 32 \\
\midrule
\multicolumn{4}{l}{\textbf{{\ourdit} details}} \\
\rowcolor{diffhl}
Size & Base & Large & XL & Base & Large & XL \\ 
\rowcolor{diffhl}
Layers & 12 & 24 & 27 & 12 & 24 & 27 \\
\rowcolor{diffhl}
Hidden dims & 768 & 1024 & 1408 & 768 & 1024 & 1408 \\
\rowcolor{diffhl}
Numbers of Heads & 12 & 16 & 22 & 12 & 16 & 22 \\
Dims per heads & 64 & 64 & 64 & 64 & 64 & 64 \\
Patch size & 1 & 1 & 1 & 1 & 1 & 1 \\
Class drop ratio & 10\% & 10\% & 10\% & 10\% & 10\% & 10\% \\

\midrule
\multicolumn{4}{l}{\textbf{VAE loss details}}\\
MSE weight & 1.0 & 1.0 & 1.0 & 1.0 & 1.0 & 1.0 \\
LPIPS weight & 1.0 & 1.0 & 1.0 & 1.0 & 1.0 & 1.0 \\
GAN weight & 0.1 & 0.1 & 0.1 & 0.1 & 0.1 & 0.1 \\
GAN starting step & 0 & 0 & 0 & 0 & 0 & 0  \\
GAN LeCam weight & 0.05 & 0.05 & 0.05 & 0.05 & 0.05 & 0.05\\
GAN LeCam EMA & 0.999 & 0.999 & 0.999 & 0.999 & 0.999 & 0.999\\
KL weight & 1e-5 & 1e-5 & 1e-5 & 1e-5 & 1e-5 & 1e-5 \\

\midrule
\multicolumn{4}{l}{\textbf{Diffusion loss details}}\\
\rowcolor{diffhl}
Time sampler & Log-norm (0.0, 1.0) & Uniform (0.0, 1.0) & Uniform (0.0, 1.0) & Log-norm (0.0, 1.0) & Uniform (0.0, 1.0) & Uniform (0.0, 1.0) \\
Loss weight & 1.0 & 1.0 & 1.0 & 1.0 & 1.0 & 1.0 \\
Scheduler & Linear & Linear & Linear & Linear & Linear & Linear\\ 

\midrule
\multicolumn{4}{l}{\textbf{Tweedie Pixel Reconstruction (TPR) loss details }} \\
L2 weight & 1.0 & 1.0 & 1.0 & 1.0 & 1.0 & 1.0 \\
LPIPS weight & 1.0 & 1.0 & 1.0 & 1.0 & 1.0 & 1.0 \\

\midrule
\multicolumn{4}{l}{\textbf{REPA++ loss details }} \\
Latent REPA weight  & 1.0 & 1.0 & 1.0 & 1.0 & 1.0 & 1.0  \\
Diffusion REPA weight & 1.0 & 1.0 & 1.0 & 1.0 & 1.0 & 1.0 \\
Diffusion REPA depth & 8 & 8 & 8 & 8 & 8 & 8 \\

\midrule
\multicolumn{4}{l}{\textbf{Data augmentation details }}\\
Center Crop & $\checkmark$ & $\checkmark$ & $\checkmark$ & $\checkmark$ & $\checkmark$ & $\checkmark$ \\
Random Horizontal Flip & $\checkmark$ & $\checkmark$ & $\checkmark$ & $\checkmark$ & $\checkmark$ & $\checkmark$ \\

\midrule
\multicolumn{4}{l}{\textbf{Training details }}\\
Batch size & 256 & 256 & 256 & 256 & 256 & 256 \\
Iterations & 1M & 1M & 1M & 1M & 1M & 1M \\
Epochs & 200 & 200 & 200 & 200 & 200 & 200 \\
Learning rate & 1e-4 & 1e-4 & 1e-4 & 1e-4 & 1e-4 & 1e-4 \\
Adam $\beta_1$ & 0.9 & 0.9 & 0.9 & 0.9 & 0.9 & 0.9 \\
Adam $\beta_2$ & 0.999 & 0.999 & 0.999 & 0.999 & 0.999 & 0.999 \\
Weight decay & 0.0 & 0.0 & 0.0 & 0.0 & 0.0 & 0.0 \\
EMA decay rate & 0.9999 & 0.9999 & 0.9999 & 0.9999 & 0.9999 & 0.9999 \\

\midrule
\multicolumn{4}{l}{\textbf{Sampling details }}\\
SDE & $\checkmark$ & $\checkmark$ & $\checkmark$ & $\checkmark$ & $\checkmark$ & $\checkmark$ \\
Solver & Euler–Maruyama & Euler–Maruyama & Euler–Maruyama & Euler–Maruyama & Euler–Maruyama & Euler–Maruyama \\
NFE & 250 & 250 & 250 & 250 & 250 & 250 \\

\bottomrule
\end{tabular}
}%
\end{table*}


\end{document}